\documentclass[10pt,twocolumn,letterpaper]{article}

\usepackage[pagenumbers]{config/cvpr}

\usepackage{amsmath}
\usepackage{booktabs}
\usepackage{amssymb}
\usepackage[dvipsnames]{xcolor}

\newcommand{\purple}[1]{#1}

\newcommand{\todo}[1]{#1}
\newcommand{\TODO}[1]{#1}

\newcommand{\lixin}[1]{\textcolor{cyan}{#1}} %

\newcommand{\nameMethod}{\mbox{{RHINO}}\xspace}
\newcommand{\nameMethodLONG}{{\textcolor{black}{R}econstructing \textcolor{black}{H}uman \textcolor{black}{I}nteractions with \textcolor{black}{N}ovel \textcolor{black}{O}bjects}\xspace}
\newcommand{\nameDatasetGT}{\mbox{{BenchRHINO}}\xspace}
\newcommand{\nameDatasetITW}{\mbox{{WildRHINO}}\xspace}

\newcommand{\SDF}{\mbox{SDF}\xspace}
\newcommand{\SDFs}{\mbox{SDFs}\xspace}

\newcommand{\nSDFs}{\mbox{neural SDFs}\xspace}

\newcommand{\zheading}[1]{\textbf{#1.}}
\newcommand{\qheading}[1]{\noindent\textbf{#1.}}

\newcommand{\smpl}{\mbox{SMPL}\xspace}
\newcommand{\SMPL}{\smpl}

\newcommand{\HOI}{\mbox{HOI}\xspace}
\newcommand{\HOIs}{\mbox{HOIs}\xspace}
\newcommand{\smplx}{\mbox{SMPL-X}\xspace}
\newcommand{\smplX}{\smplx}

\newcommand{\sota}[0]{\mbox{SotA}\xspace}

\newcommand{\supmat}{\textcolor{black}{{{Supp.~Mat.}}}\xspace}

\usepackage{lipsum}
\usepackage{pifont}

\usepackage{balance}

\usepackage{tikz}
\usepackage{graphicx}
\usepackage{overpic}
\usepackage{multirow}

\usepackage{pifont}

\usepackage{fontawesome}

\newcommand{\cg}[1]{{\color{magenta}[CG: #1]}}

\usepackage[normalem]{ulem}

\usepackage{enumitem}

\renewcommand{\todo}[1]{}
\renewcommand{\TODO}[1]{}
\renewcommand{\lixin}[1]{}
\renewcommand{\cg}[1]{}

\definecolor{brickred}{rgb}{0.8, 0.25, 0.33}
\definecolor{brightmaroon}{rgb}{0.76, 0.13, 0.28}
\definecolor{brightlavender}{rgb}{0.75, 0.58, 0.89}
\definecolor{byzantine}{rgb}{0.74, 0.2, 0.64}
\definecolor{byzantium}{rgb}{0.44, 0.16, 0.39}

\newcommand{\chore}[0]{\mbox{CHORE}\xspace}

\newcommand{\procigen}[0]{\mbox{ProciGen}\xspace}

\newcommand{\HSR}{\mbox{HSR}\xspace}

\newcommand{\HOLD}{\mbox{HOLD}\xspace}

\newcommand{\HOLDplusRHINO}{\mbox{HOLD$^*$}\xspace}

\newcommand{\SFM}{\mbox{SfM}\xspace}

\newcommand{\MASTER}{\mbox{MASt3R}\xspace}

\newcommand{\SAM}{\mbox{SAM2}\xspace}

\newcommand{\RANSAC}{\mbox{RANSAC}\xspace}
\newcommand{\ransac}{\RANSAC}
\newcommand{\InteractVLM}{\mbox{InteractVLM}\xspace}

\newcommand{\InterTrack}{\mbox{InterTrack}\xspace}

\newcommand{\aios}{\mbox{AiOS}\xspace}
\newcommand{\sapiens}{\mbox{Sapiens}\xspace}

\newcommand{\loftr}{\mbox{LoFTR}\xspace}
\newcommand{\LOFTR}{\loftr}
\newcommand{\superpoint}{\mbox{SuperPoint}\xspace}
\newcommand{\SuperPoint}{\superpoint}

\newcommand{\superglue}{\mbox{SuperGlue}\xspace}
\newcommand{\SuperGlue}{\superglue}
\newcommand{\RGB}{\mbox{RGB}\xspace}

\newcommand{\scene}{\text{scn}}
\newcommand{\object}{\text{obj}}

\newcommand{\transfHomogScale}{\mathbf{S}}
\newcommand{\transfScale}{\mathbf{s}}
\newcommand{\transfHomogRigid}{\mathbf{T}}
\newcommand{\transfHomogRot}{\mathbf{R}}
\newcommand{\transfHomogTransl}{\mathbf{t}}

\newcommand{\Cam}{\mathbf{C}}
\newcommand{\cam}{\mathbf{c}}
\newcommand{\CamScene}{\Cam_\scene}

\newcommand{\camScenei}{\cam_\scene^i}
\newcommand{\CamObject}{\Cam_\object}

\newcommand{\camObjecti}{\cam_\object^i}

\newcommand{\poses}{\mathbf{P}}

\newcommand{\mycmark}{\textcolor{green!60!black}{\ding{51}}} %
\newcommand{\myxmark}{\textcolor{red}{\ding{55}}} %

\newcommand\mymapsto{\mathrel{\ooalign{$\rightarrow$\cr\kern-.15ex\raise.275ex\hbox{\scalebox{1}[0.522]{$\mid$}}\cr}}}

\definecolor{cvprblue}{rgb}{0.21,0.49,0.74}
\usepackage[pagebackref,breaklinks,colorlinks,allcolors=cvprblue]{hyperref}
\usepackage[accsupp]{axessibility}

\usepackage[capitalize]{cleveref}
\crefname{figure}{Fig.}{Figs.}
\Crefname{figure}{Figure}{Figures}
\crefname{section}{Sec.}{Secs.}
\Crefname{section}{Section}{Sections}
\Crefname{table}{Table}{Tables}
\crefname{table}{Tab.}{Tabs.}
\Crefname{equation}{Equation}{Equations}
\crefname{equation}{Eq.}{Eqs.}

\def\paperID{*****}
\def\confName{CVPR}
\def\confYear{2026}
\usepackage{xcolor,colortbl}

\title{\vspace{-0.75 em}
\textcolor{cvprblue}{RHINO}:
\textcolor{cvprblue}{R}econstructing
\textcolor{cvprblue}{H}uman
\textcolor{cvprblue}{I}nteractions with
\textcolor{cvprblue}{N}ovel
Objects   \\
from Monocular Videos\vspace{-0.50 em}}

\author{
Lixin Xue$^1$               \quad %
Chengwei Zheng$^{1,2}$         \quad %
Georgios Paschalidis$^3$    \quad %
\\
Chen Guo$^1$                \quad %
Manuel Kaufmann$^1$         \quad %
Juan Zarate$^1$             \quad %
Dimitrios Tzionas$^{3,4}$
\vspace{+0.5 em}
\\
\vspace{+0.5 em}
{
\small
$^1$ETH Zürich \quad
$^2$The University of Tokyo \quad
$^3$University of Amsterdam \quad
$^4$Aristotle University of Thessaloniki \quad
}
}

\begin{document}

\twocolumn[{%
\renewcommand\twocolumn[1][]{#1}%
\maketitle
\begin{center}
    \centering
    \vspace{-2.2 em}
    \phantomsection
    \includegraphics[width=0.99 \textwidth]{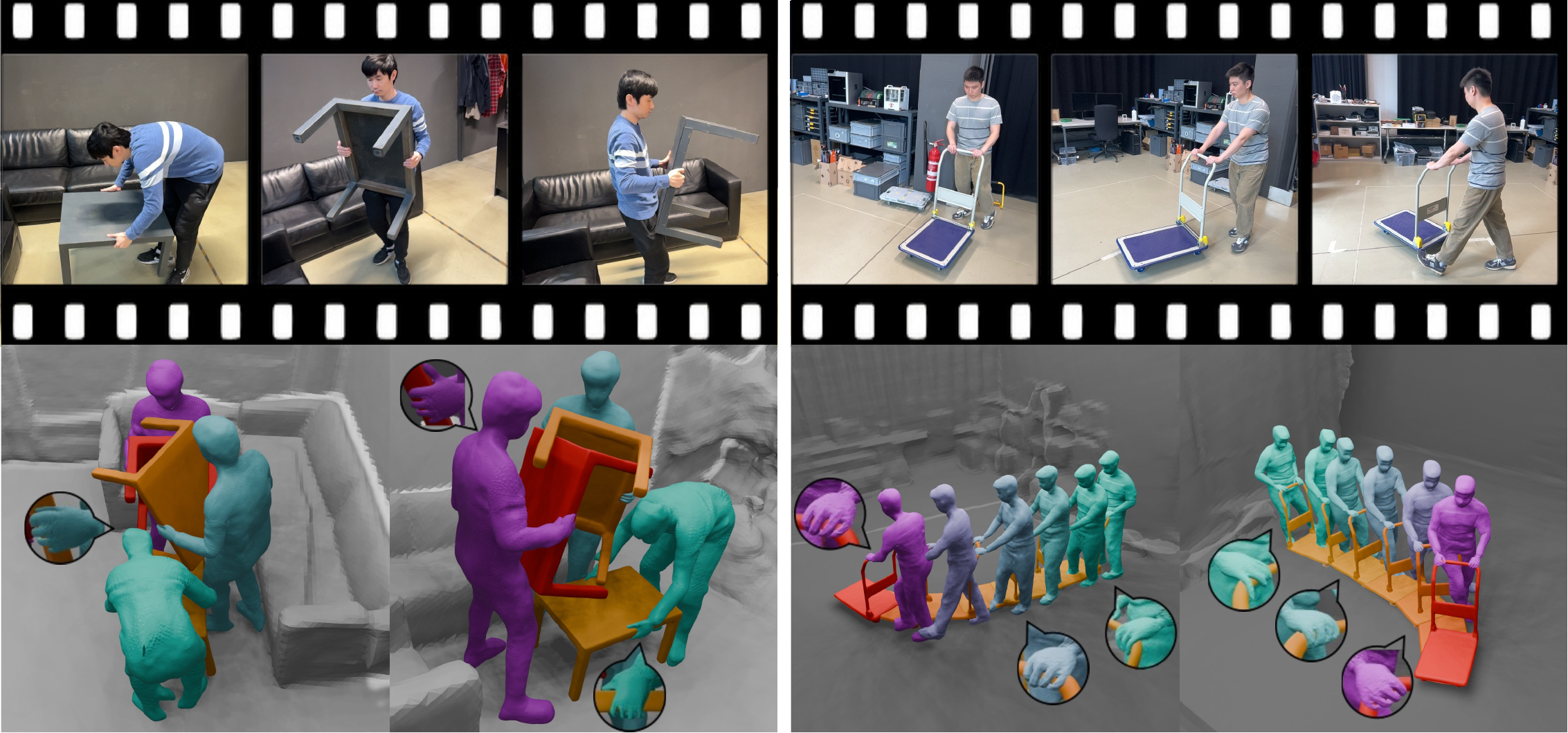}
    \vspace{-0.7em}
    \captionof{figure}{
        We develop \nameMethod, a novel framework that reconstructs detailed (dynamic) 3D human-object interactions (\HOI) and the surrounding scene within a common world frame %
        from a monocular \RGB video with a moving viewpoint. 
        \nameMethod uses per-component \nSDFs to: %
        (i)  capture shape details, and 
        (ii) encourage contact via a differentiable distance term. 
        The ``zoom-in insets'' highlight plausible contacts. 
        \nameMethod requires neither a pre-scanned object template nor prior knowledge of the object, unlike most existing work. 
    }
    \label{fig:teaser}
    \vspace{+0.72 em}
\end{center}

}]

\begin{abstract}
Reconstructing people, objects, and their interactions in 3D is a long-standing and fundamental %
goal for intelligent systems. 
Often the input is \RGB video from a %
moving camera, making the task ill-posed; depth is ambiguous, 
humans and objects occlude each other, 
and camera and object motion entangle to create apparent motion. 
Most prior work addresses humans or objects in isolation, 
ignoring their interplay, 
or assumes known 3D shapes or cameras, which is impractical for real-world applications. 
We develop \nameMethod (\nameMethodLONG), a novel three-step framework that 
recovers in 3D a human, \purple{novel (unseen)} manipulated object, and static scene %
in a common world frame from a monocular \RGB video.
First, we leverage 3D-aware foundation models to obtain cues that stabilize Structure-from-Motion (SfM) even for low-texture regions; this yields a coarse shape and \mbox{apparent} motion of 
a manipulated object from foreground pixels, and a coarse scene shape and camera motion from background pixels. 
Second, we estimate a human in the camera frame via an off-the-shelf method, and 
subtract the camera motion from apparent motion to extract the object motion; 
this registers the human, object, and coarse scene shapes into a common world frame. 
Third, we refine shapes using a compositional neural field with per-component signed-distance fields.
The latter further enables differentiable contact priors that attract surfaces while penalizing interpenetration, improving the physical plausibility of the final reconstruction.
For evaluation, we capture a new dataset of handheld monocular videos synchronized with a volumetric 4D capture stage, providing ground-truth shape and camera motion. 
\nameMethod outperforms state-of-the-art baselines on novel-view synthesis and 4D reconstruction. 
Ablations show that each stage contributes substantially. 
Code and data are available 
at 
\href{https://lxxue.github.io/RHINO}{https://lxxue.github.io/RHINO}.
\enlargethispage{1.5 em}
\end{abstract}

\section{Introduction}
\label{sec:intro}

Humans constantly interact with and manipulate objects in their surroundings. 
Enabling computers to perceive these interactions in 3D from a single RGB video 
benefits assistive robotics, AR/VR, healthcare, media, and learning from internet-scale videos. 
In this work, we reconstruct a 4D scene with dynamic human-object interactions (\HOI) from such a monocular video.

This task is challenging due to depth ambiguities and human-object occlusions. 
Moreover, camera motion produces ``apparent motion'' that entangles camera and object motion. 
Finally, while 3D human recovery is now tractable, reconstructing manipulated objects remains highly challenging because objects vary widely in shape and appearance and are often low-texture or symmetric, making feature detection and tracking difficult.

\begin{figure}
    \centering
        \vspace{-0.5 em}
    \captionsetup{type=figure}%
        \includegraphics[width=0.99 \columnwidth]{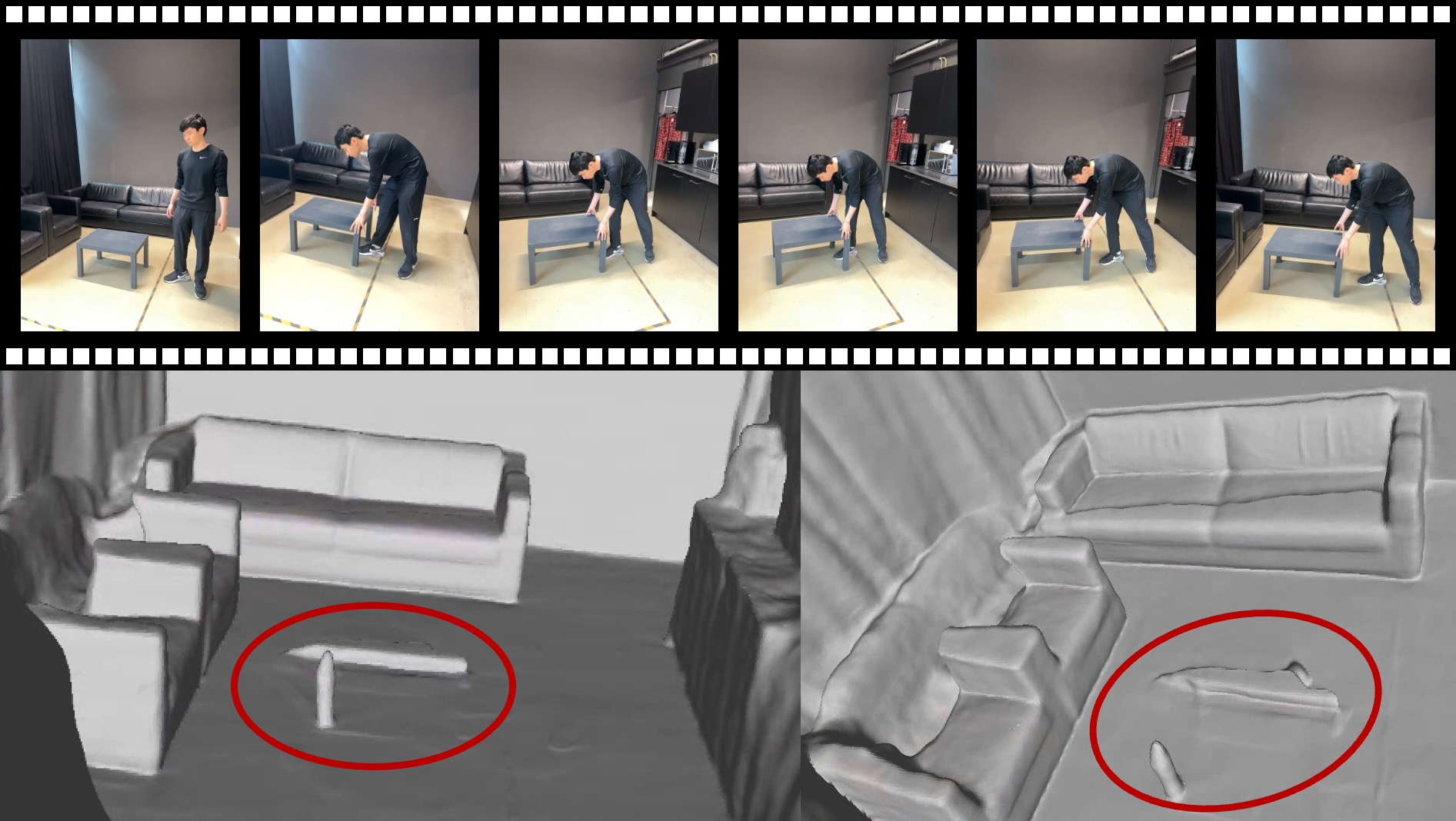}
        \vspace{-0.5 em}
    \caption{
        \textbf{Existing work}, such as the \HSR \cite{xue2024hsr} \sota method, can faithfully reconstruct the 3D shape of a static scene and of a person %
        moving in it, 
        but struggles when people manipulate objects. 
        As illustrated, %
        when a person pushes a table (top), the static part of the scene is reconstructed well (bottom row, two views), but the table's reconstruction is degenerate 
        (see the red highlight). 
        Here we do not show the reconstructed person to reduce occlusions.
    }
    \label{fig:sota_fails}
    \vspace{-0.5 em}
\end{figure}

Due to these challenges, prior methods address only parts of the problem. 
Most methods reconstruct human-free scenes \cite{teed2021droid, dust3r, wang2025vggt, chen2025back} or 
humans in isolation \cite{guo2023vid2avatar, guo2025vid2avatarpro, qian20233dgsavatar, poco}.  
A few methods jointly reconstruct humans and scenes in a common world frame, but still fail to capture manipulation-induced object motion \cite{xue2024hsr,zhang2025odhsr}; see \cref{fig:sota_fails}. 
Moreover, many methods assume known object/scene shapes \cite{hassan2019resolving, bhatnagar2022behave, guzov2021human,huang2024intercap} or calibrated cameras \cite{huang2022capturing, bhatnagar2022behave}, 
which is often impractical.

To our knowledge, no method provides %
a framework for recovering a 4D HOI scene in a world frame from a moving-view monocular \RGB video.
We address this gap with \nameMethod (\emph{\nameMethodLONG}), a three-stage framework; see its results in \cref{fig:teaser}.

First, we estimate the shape and motion of a \purple{novel (unseen)} object. 
Prior work~\cite{sun2022onepose, he2022oneposeplusplus} tackles this via Structure-from-Motion (\SFM) and feature correspondences. 
However, everyday objects in full‑body videos are often low‑texture and occupy a small region of the frame, making sparse features \cite{detone18superpoint} unstable and dense correspondences \cite{sun2021loftr} inconsistent across frames.
We leverage recent 3D‑aware foundation models \cite{mast3r} to produce dense, robust correspondences that enable more reliable \SFM for these objects.

However, the camera also moves, so apparent motion entangles camera and object motion.
To disentangle the two motions, we first estimate the camera motion from static background regions via \SFM, and scale and align it with the apparent motion.
Then, we ``subtract'' this camera‑induced component from the apparent motion to obtain object motion.
We also estimate an initial human shape and motion in the camera frame \cite{sun2024aios}.
Overall, this stage registers the human, object, and scene into a common world frame.

The initial human, object, and scene shapes are coarse.
We refine them using a compositional neural field with per-component signed-distance and appearance fields.
We optimize the fields for photometric and mask consistency, with geometric regularization.
This yields detailed 3D human, object, and scene shapes aligned with the image cues.
Crucially, operating in a world frame—rather than individual camera frames—enables multi-frame optimization over all components, mitigating per-frame initialization errors.

However, reconstructed 3D shapes often lack physically-plausible contact; hands frequently hover above objects or visibly interpenetrate them.
Our key insight is that neural signed distance fields (\nSDFs) not only encode geometry, but also provide continuous, differentiable distances to object surfaces—an ideal signal for reasoning about contact. 
We therefore repurpose per-component \nSDFs to formulate a contact-aware loss that simultaneously attracts surfaces and penalizes inter-penetrations, improving physical plausibility even under occlusions. 
We show that using \nSDFs to jointly encode shape and reason about contact is effective for monocular \HOI reconstruction.

For evaluation, we capture a \purple{new dataset of 7 sequences} 
using a handheld camera synchronized with a 4D capture stage to obtain monocular \RGB video paired with ground-truth 4D \HOI geometry and camera motion. 
We evaluate on novel-view synthesis and 4D reconstruction; 
\nameMethod outperforms state-of-the-art (\sota) baselines. %
Ablations show that each of \nameMethod's three stages contributes significantly.

Overall, our main contributions are the following:
\begin{enumerate}[topsep=0pt,noitemsep, nolistsep] %
    \setlength\itemsep{0.2 em}
    \item
    We address the problem of reconstructing a 4D \HOI scene in a world frame from monocular \RGB video, a setting not tackled by prior work.
    \item
    We use a compositional neural field framework with per‑component appearance fields and \SDFs; the latter encode both shape and human-object contact cues. 
    \item
    We use 3D-aware foundation models to estimate object motion, while decoupling it from camera motion. 
    \item
    We collect a new evaluation dataset with a handheld, moving \RGB camera paired with 4D ground truth. 
\end{enumerate}
Code and data are public 
at 
\href{https://lxxue.github.io/RHINO}{https://lxxue.github.io/RHINO}.

\section{Related Work}

\hspace{\parindent}
\zheading{Object Pose Estimation}
Template-based pose estimation methods~\cite{moon2024genflow, caraffa2024freeze, Di_2021_ICCV, Lin_2024_CVPR, wang2019normalized} achieve strong performance by tracking a known object template, but the need for a pre-acquired template limits their applicability in the wild.
Template-free methods~\cite{lee2025any6d, antic2025sdfit} avoid this assumption, yet current approaches still require depth input~\cite{wen2024foundationpose} or a static reference video to build an object model~\cite{sun2022onepose, he2022oneposeplusplus}, limiting their use in the wild.
Furthermore, these approaches rely on traditional 2D feature-matching techniques~\cite{detone18superpoint, sun2021loftr} that lack multi-view consistency and often produce unreliable correspondences.
In contrast, we leverage 3D foundation models~\cite{mast3r} for dense, geometrically consistent correspondences, enabling robust pose initialization without pre-scanned templates. We further reconstruct objects in 3D, and refine object poses using photometric cues, achieving accurate pose estimation and high-fidelity reconstruction.

\zheading{3D Human-Object Reconstruction}
Most work 
on 3D %
\HOI %
relies on object templates that are known~\cite{xie2022chore, xie2025intertrack, nam2024joint, weng2021holistic, guo2020single, zhao2024imhoi, kjellstrom2010trackingPeopleObjects} or \purple{retrieved from databases \cite{cseke_tripathi_2025_pico, dwivedi_interactvlm_2025, zhang2020phosa}}, limiting applicability.
\procigen~\cite{xie2024template}, trained on a large synthetic dataset, enables template-free 
\HOI 
reconstruction.
\mbox{InterTrack}~\cite{xie2025intertrack} 
extends this 
to coherently track 
\HOI in 3D 
over videos.
Most existing methods are evaluated on datasets such as \mbox{BEHAVE}~\cite{bhatnagar2022behave} and \mbox{InterCap}~\cite{huang2024intercap}, focusing on minimally clothed bodies—neglecting detailed surface geometry that serves as a critical contact cue (\eg, shoes).
Recent methods~\cite{gopal2025betsu, jiang2022neuralhofusion, zhang2023neuraldome, zhang2024hoi} %
reconstruct detailed interacting humans and objects with separable representations, but 
rely on 
multi-view input, which is rarely available in practice.
\purple{Hand-only tracking and contact cues have been used to reconstruct unseen objects \cite{tzionas2015inhand, panteleris2015inhand}, but interactions are simple.} 
Recently, \mbox{HOLD}~\cite{fan2024hold} attains detailed hand–object reconstructions from a single video, yet it 
does not tackle 
full-body interactions. %
Unlike \chore and \InterTrack, which operate in the camera frame, \nameMethod reconstructs full-body human–object interactions in a world frame, including the scene, directly from monocular RGB video—without requiring 
prior object knowledge.

\zheading{3D Human-Scene Reconstruction}
Existing methods 
recover 3D 
human 
\purple{skeletons \cite{savva2016pigraphs, chen2019holistic++, gupta2011workspace, iMapper2018} or} 
meshes \cite{hassan2019resolving, huang2022capturing, guzov2021human, zhang2022egobody, dai2022hsc4d, kaufmann2023emdb, muller2025reconstructing, liu2025joint, SceneAware_EG2023, Li_3DV2022, yi2022mover} within static scenes.
However, these methods typically rely on pre-scanned scene geometry.
Recent feedforward models~\cite{chen2025human3r, rojas2025hamst3rhumanawaremultiviewstereo} relax this dependency by inferring camera and human parameters from images, though at reduced accuracy and without explicit human modeling. 
Several works~\cite{song2023total, yang2023ppr, xue2024hsr, zhang2025odhsr} seek joint human-scene reconstruction from monocular inputs. 
\mbox{HSR}~\cite{xue2024hsr} reconstructs humans within static scenes but cannot handle moving objects.
Going further, \nameMethod tackles the challenging task of additionally reconstructing the dynamically manipulated object—whose shape, pose, and motion are all unknown—in the same world frame.

\section{Method}    \label{sec:method}
    We consider an input video containing a human interacting with an object, captured with a single, moving \RGB camera. 
    Our goal is to recover the detailed 3D shapes and appearances of a human, a manipulated object, and their surrounding environment in a 
    common world frame.
    
\begin{figure*}
    \centering
    \vspace{-0.5 em}
    \captionsetup{type=figure}%
        \includegraphics[width=0.99 \textwidth, trim={0 0 0 0}, clip]{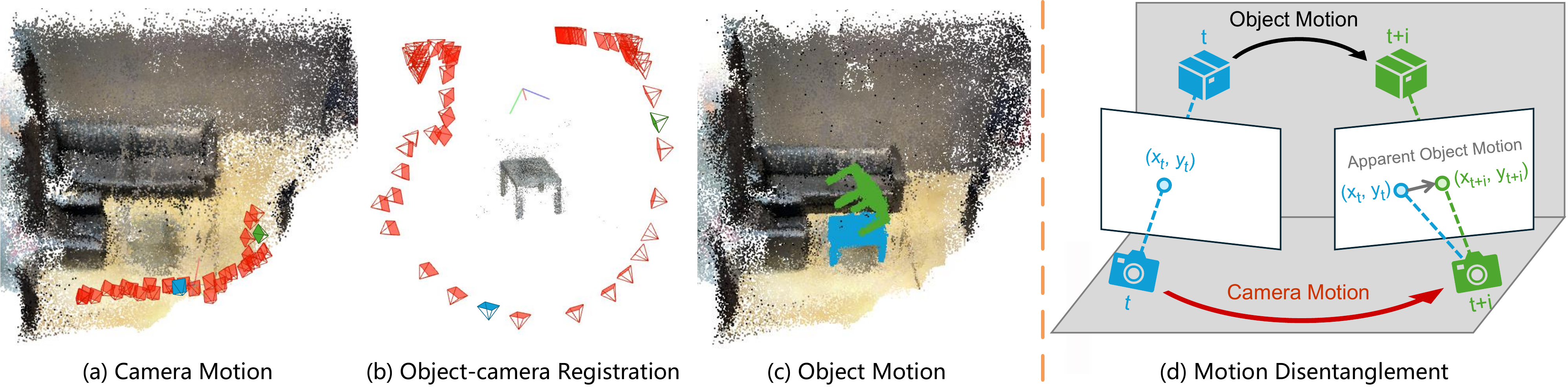}
    \vspace{-0.5 em}
    \caption{
                    \textbf{Camera \& Object motion (\cref{sec:method-initialization}, \cref{sec:method-frame-alignment}}). 
                    When both the camera and object move, their motion entangles into ``apparent'' motion.
                    To disentangle them, we 
                    (a) estimate camera motion in the world frame via \SFM on scene-only pixels, 
                    (b) estimate apparent motion via \SFM on object-only pixels, 
                    (c--d) estimate object motion in the world frame by ``removing'' the camera motion from the apparent one. %
    }
    \label{fig:motion_decomposition}
\end{figure*}

\begin{figure*}
    \centering
    \vspace{-1.5 em}
    \captionsetup{type=figure}%
        \includegraphics[width=0.99 \textwidth, trim={0 2em 0 0}, clip]%
        {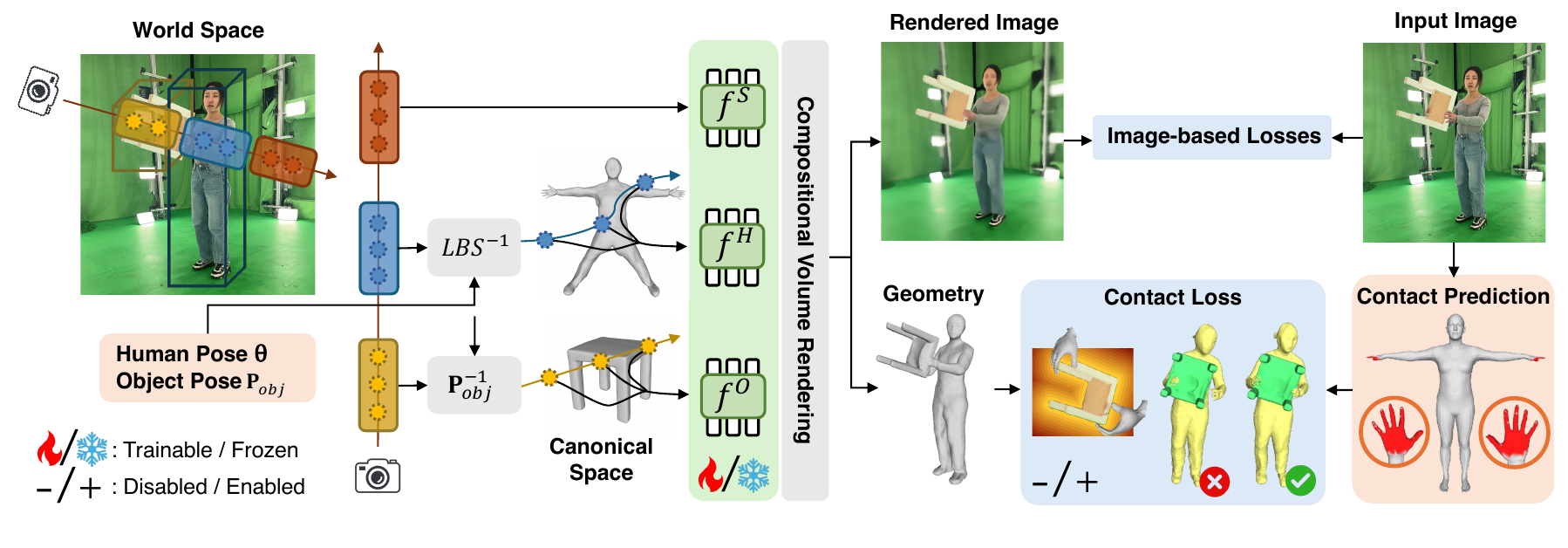}
    \vspace{-0.5 em}
    \caption{
        \textbf{Method Overview.}
        Starting with initialized global human and decoupled object poses (\cref{sec:method-initialization}, \cref{sec:method-frame-alignment}), we sample points along the camera ray for the human, object and static scene. 
        To enable consistent representation, sampled points are warped into canonical space using inverse LBS for the human and the estimated rigid transformation for the object.
        All components are then rendered holistically via compositional volume rendering. 
        A global optimization (\cref{sec:method-refinement-SDFs}) 
        helps 
        learn the 3D representation of all elements and refine the initial poses via a photometric loss, while encouraging physically plausible contact by leveraging differentiable contact priors (\cref{sec:method-contact}).} 
        \vspace{-0.5 em}
    \label{fig:arch}
\end{figure*}

    We build a three-stage framework  %
    (\cref{fig:motion_decomposition,fig:arch}).
    First, %
    we estimate
    an initial scene, object, and human (\cref{sec:method-initialization}), and %
    align them into a common world frame (\cref{sec:method-frame-alignment}).
    Then, we recover details via compositional per-component \nSDFs
    (\cref{sec:method-refinement-SDFs}).
    Lastly, we use these \nSDFs to
    encourage contact and avoid inter-penetration
    (\cref{sec:method-contact}).

\subsection{Initialization}
\label{sec:method-initialization}
    We estimate coarse shape for the static scene, and coarse shape and motion for the moving object and human 
    in the camera frame. 
    We also estimate %
    camera motion. 
    This lets us later 
    align  %
    the human, object, and scene into a common world frame (\cref{sec:method-frame-alignment}), %
    and jointly refine these %
    (\cref{sec:method-refinement-SDFs}).

    \zheading{Camera Motion \& Scene Initialization} %
        We estimate the camera motion under the assumption that the scene background is static and defined in the world frame. %
        Consequently, any apparent motion between the scene and the camera is caused only by the camera motion. 
        However, the input video also contains the dynamic human and object. 
        To isolate the static background, we exploit \SAM \cite{ravi2025sam2} to segment the video into a scene-only video, masking out the human and the object. 
        Based on these background pixels, we perform \SFM~\cite{mast3r, pan2024glomap} and estimate camera motion for all frames $C_\scene = \{ C_\scene^i \}_{i=1}^N$ where $C_\scene^i$ is the camera pose (extrinsics) for frame $i$ and $N$ is the video length, and a rough 3D scene point cloud, $PCL_\scene$, as shown in \cref{fig:motion_decomposition}~\textcolor{cvprblue}{(a)}.

    \zheading{Object Pose Initialization}
        We segment the moving object from the video and establish feature correspondences across adjacent frames on the object pixels.
        However, the object usually occupies a small area in images, and it might be occluded, low-texture, or symmetric. 
        This makes standard keypoint detectors (\eg, \superpoint~\cite{detone18superpoint, fan2024hold}) or keypoint-free feature matching (\eg, \loftr~\cite{sun2021loftr, he2022oneposeplusplus}) yield sparse or non-distinctive matches, which challenges Structure-from-Motion (\SFM). 
        We empirically find that recent 3D-aware foundational models remain surprisingly effective. 
        Thus, we adopt \MASTER~\cite{mast3r} to establish correspondences across neighboring frames. 
        \MASTER casts this as a 3D task based on pointmap regression rather than a 2D problem in image space, resulting in feature correspondences that are more accurate and robust, even in low-texture regions.
        Based on these reliable matches, we perform triangulation and obtain %
        a composed camera trajectory, $C_\object$, as if the object was static, as shown in \cref{fig:motion_decomposition}~\textcolor{cvprblue}{(b)}.

    \zheading{Human Initialization}
        We infer \smplX~\cite{SMPL-X:2019} bodies in the camera frame via the \aios~\cite{sun2024aios} model applied on a per-frame basis.
        These bodies have a reasonable but rough pose, so their pixel-alignment can be improved. 
        Thus, we refine poses to align to 
        \sapiens \cite{khirodkar2024sapiens} 2D keypoints, 
        while regularizing poses with a temporal smoothness loss, 
        and interpolating keypoints in the case of %
        occlusions by 
        objects.

    \subsection{Aligning into World Frame}
    \label{sec:method-frame-alignment}

        \Cref{sec:method-initialization} defines two camera trajectories, 
        $C_\scene$ and 
        $C_\object$,  
        in the scene (or world) and object frame, respectively. %
        But in reality we have one camera, 
        so we need to ``unify'' these. 
        
        \zheading{Object to World}
            We consider as real the camera trajectory estimated from the static scene, $\CamScene$, 
            so we need to align $C_\object$ to $C_\scene$. %
            Let 
            $\transfHomogScale$ be a scaling and 
            $\transfHomogRigid = [ \transfHomogRot,\transfHomogTransl; \mathbf{0}, 1 ]$ a rigid transformation, and
            $\poses_\object = \{ \poses_\object^i \}_{i=1}^N$ be the object motion \purple{(sequence of poses)} 
            in the world frame.
            Then:
            \begin{equation}
                \small
                \transfHomogRigid   \cdot \transfHomogScale  \cdot  \CamObject  =  \CamScene  \cdot  \poses_\object
                \text{,}
                \label{eq:camera_transform_with_pose}
            \end{equation}
            where we need to estimate 
            $\transfHomogScale, \transfHomogRigid$.
            We identify time frames $i'$ where the object is static, using \RANSAC to find a similarity transform between $C_\object$ and $C_\scene$;
            the frames that provide consensus are static-object frames.
            For these frames,
            $P_{obj}^{i'} = I$, so \cref{eq:camera_transform_with_pose} simplifies to:
            \begin{equation}
                \small
                \transfHomogRigid  \cdot  \transfHomogScale  \cdot  \CamObject  =  \CamScene
                \text{,}
                \label{eq:camera_transform_without_pose}
            \end{equation}
            which helps solve for scale, $\transfHomogScale$, %
            rotation, $\transfHomogRot$, and translation, $\transfHomogTransl$, through the Umeyama least-squares algorithm~\cite{umeyama1991estimation}: 
            \begin{equation}
                \small
                \min_{\transfScale,\transfHomogRot,\transfHomogTransl} 
                {\sum}_{i'=1}^{n} 
                \| 
                    \transfScale \transfHomogRot \camObjecti + \transfHomogTransl - \camScenei 
                \|^2
                \text{,}
            \end{equation}
            where $\camObjecti, \camScenei$ are the camera centers of $\CamObject, \CamScene$. 
            We then solve for object pose in the world frame (\cref{fig:motion_decomposition}~\textcolor{cvprblue}{(c--d)}):%
            \begin{equation}
                \small
                \poses_\object  =  \CamScene^{-1}  \cdot   \transfHomogRigid  \cdot   \transfHomogScale  \cdot \CamObject 
                \text{,}
            \end{equation}
            by 
            removing the real-camera motion, $\CamScene$, from the apparent motion, 
            $\CamObject$, after morphing $\CamObject$ via Procrustes ($\transfHomogScale, \transfHomogRigid$). %

        \zheading{Human to World}
            The per-frame bodies of \cref{sec:method-initialization} live %
            in the camera frame, under a weak-perspective assumption.
            We recover the body's trajectory in the world frame
            under a perspective camera by exploiting 2D projection constraints along with
            3D contact constraints with a ground estimated by applying \ransac on $PCL_\scene$, as in \cite{xue2024hsr,jiang2022neuman}.

\subsection{Joint Optimization}
\label{sec:method-refinement-SDFs}

    Given initial scene, object, and human estimates
    in a common world frame (\cref{sec:method-frame-alignment}), we refine these to recover details
    via a compositional neural field that has per-component neural Signed-Distance Fields (\SDF) and appearance fields.
    Joint optimization in a shared world frame enforces multi-view consistency and mitigates initialization imperfections.

    \zheading{3D Shape Representation}
    Extending the modeling paradigm in \cite{guo2023vid2avatar,xue2024hsr}, we represent the shape of a human ($H$), object $(O)$, and scene ($S$) as \nSDFs:
        \begin{align}
            f_{\text{sdf}}^{H}: \mathbb{R}^{3 + n_{\theta_b}} & \rightarrow \mathbb{R}^{1 + n_{z}};   \  (\mathbf{x}^H, \boldsymbol{\theta}_b) \mymapsto (\xi^H, \mathbf{z}^H)    \label{eq:sdf_human}  \text{,} \\
            f_{\text{sdf}}^{O}: \mathbb{R}^{3} & \rightarrow \mathbb{R}^{1 + n_{z}};                \  \mathbf{x}^O  \mymapsto (\xi^O, \mathbf{z}^O)                          \label{eq:sdf_object} \text{,} \\
            f_{\text{sdf}}^{S}: \mathbb{R}^{3} & \rightarrow \mathbb{R}^{1 + n_{z}};                \  \mathbf{x}^S  \mymapsto (\xi^S, \mathbf{z}^S)                          \label{eq:sdf_scene}  \text{.}
        \end{align}
        where $f_{\text{sdf}}^{(\cdot)}$ maps a 3D point $\mathbf{x}^{(\cdot)}$ to a signed distance value, $\xi^{(\cdot)}$, and a geometric feature, $\mathbf{z}^{(\cdot)}$.
        To capture pose-dependent deformations (\eg, clothing wrinkles), $f_{\text{sdf}}^{H}$ conditions on body articulation $\boldsymbol{\theta}_b$\purple{, excluding global orientation and translation}.
        The human and object fields operate in canonical space; 
        for mapping to the world frame see below.

    \zheading{Appearance Representation}
    To model the appearance of a human ($H$), object $(O)$, and scene ($S$) we employ three neural fields that predict \RGB colors from 3D points:
        \begin{align}
            f_{\text{rgb}}^{H}: \mathbb{R}^{3 + n_{\theta_b} + n_{z} + 3}  & \rightarrow \mathbb{R}^{3};  \ (\mathbf{x}^H, \boldsymbol{\theta}_b, \mathbf{z}^H, \mathbf{n}^H) \mymapsto \cam^H  \label{eq:rgb_human}    \text{,} \\
            f_{\text{rgb}}^{O}: \mathbb{R}^{3 + 3 + n_{z} + 3} & \rightarrow \mathbb{R}^{3};            \ (\mathbf{x}^O, \mathbf{v}, \mathbf{z}^O, \mathbf{n}^O) \mymapsto \cam^O           \label{eq:rgb_object}   \text{,} \\
            f_{\text{rgb}}^{S}: \mathbb{R}^{3 + 3 + n_{z} + 3} & \rightarrow \mathbb{R}^{3};            \ (\mathbf{x}^S, \mathbf{v}, \mathbf{z}^S, \mathbf{n}^S) \mymapsto \cam^S           \label{eq:rgb_scene}    \text{.}
        \end{align}
    All color fields condition on shape features, $\mathbf{z}^{(\cdot)}$, and normals, $\mathbf{n}^{(\cdot)}$. 
    The latter encourages disentanglement of shape and appearance, and is obtained by computing the gradient of the respective \SDF. 
    Note that $f_{\text{rgb}}^{H}$ conditions on $\boldsymbol{\theta}_b$, while $f_{\text{rgb}}^{O}$ and $f_{\text{rgb}}^{S}$ condition on the viewing direction, $\mathbf{v}$. 
    For the object and the scene model, we optimize a per-frame latent code to capture effects like shadows and highlights.

    \zheading{Mapping Canonical to World Frame}
    We model the human and object in respective canonical, pose-independent frames and find maps from these to the world frame. 
    For the human field, we use skeletal deformations and linear blend skinning (LBS) with bone transformations extracted from $\boldsymbol{\theta}$ as in \cite{guo2023vid2avatar,xue2024hsr}. 
    To do so:
    $\mathbf{x}'^H = LBS(\mathbf{x}^H, \boldsymbol{\theta})$ and
    $\mathbf{x}^H = LBS^{-1}(\mathbf{x}'^H, \boldsymbol{\theta})$, where $\mathbf{x}'$ lives in the world space.
    For the object, we simply apply the object's pose to move between the spaces:
            $\mathbf{x}'^O = \poses_{obj} \mathbf{x}^O$ and $\mathbf{x}^O = \poses^{-1}_{obj} \mathbf{x}'^O$.

    \zheading{Compositional Volume Rendering}
        We perform a compositional volume rendering for each camera ray, $r$, to render 
        an image 
        and 
        model occlusions between %
        scene elements. %
        Specifically, we sample $N$ 3D points within the bounding box of each component in the world frame, 
        and then sort based on their distance to the camera.
        The %
        color value is: %
        \begin{equation}
            C(r) ={\sum}_{i=1}^{3N} \tau_{i} \cam^{(\cdot)}(\mathbf{x}^{i})
            \text{,}
        \end{equation}
        where $\tau_{i}$ is an opacity value defined in \cite{xue2024hsr}.
        Similarly, we can render depth, surface normals, and masks.

    \zheading{Joint Optimization}
        We train the human, scene, and object neural fields jointly to recover details, via global optimization over all $K$ frames using a per-pixel \RGB loss and auxiliary losses (mask, depth, and normal losses) as in \cite{xue2024hsr}.
        
        In our work, we model close human-object interactions, which often feature severe mutual occlusions (so reconstructed bodies seem truncated), and require detailed hand reconstruction.
        To address this, we sample points within \smplX~\cite{SMPL-X:2019} human bodies (with negative \SDF values) and penalize for points outside these (with positive \SDF).
        Due to hand dexterity and inaccurate hand pose estimation, naive reconstruction often leads to clumpy hand geometry. 
        We resolve this via a hand-specific \SDF loss that is guided by the \smplX body mesh.
        For details, see \supmat
    
    \subsection{Improving Physical Plausibility}
    \label{sec:method-contact}

    The recovered scene, object, and human align well to pixels (\cref{sec:method-refinement-SDFs}), but may be misaligned \wrt each other due to depth ambiguity; \eg, hands might hover over or penetrate an object.
    We tackle this via contact and collision losses with a two-stage alternating refinement framework.

    \newcommand{\HHuman}{H}
\newcommand{\OObject}{O}
\newcommand{\BBoth}{H+O}
\newcommand{\SScene}{S}

\begin{table*}
    \centering
    \vspace{-0.5 em}
    \begin{minipage}{0.68\textwidth}
        \centering
        \resizebox{\textwidth}{!}{
    \begin{tabular}{l ccc ccc ccc ccc}
        \toprule
        & \multicolumn{3}{c}{Setup} & \multicolumn{3}{c}{Chamfer} & \multicolumn{3}{c}{Hausdorff} & \multicolumn{3}{c}{F1 Score}              \\ 
        & \multicolumn{3}{c}{} & \multicolumn{3}{c}{Distance~[cm] $\downarrow$} & \multicolumn{3}{c}{Distance~[cm] $\downarrow$} & \multicolumn{3}{c}{@2 cm [\%] $\uparrow$} \\ 
        \cmidrule(lr){2-4} \cmidrule(lr){5-7} \cmidrule(lr){8-10} \cmidrule(lr){11-13}
        Method & \HHuman & \OObject & \SScene & 
                 \HHuman & \OObject & \BBoth & 
                 \HHuman & \OObject & \BBoth & 
                 \HHuman & \OObject & \BBoth \\ \midrule
        \HSR~\cite{xue2024hsr} & \mycmark & \myxmark & \mycmark & 2.69 & --- & --- & 22.28 & --- & --- & 55.17 & --- & --- \\ 
        \HOLD~\cite{fan2024hold}  & \myxmark & \mycmark & \myxmark & --- & 4.41 & --- & --- & 11.92 & --- & --- & 33.64 & --- \\ 
        \InterTrack~\cite{xie2025intertrack} & \mycmark & \mycmark & \myxmark & 4.66 & 11.16 & 7.18 & 20.58 & 33.28 & 30.97 & 29.41 & 16.81 & 25.02 \\ \midrule
        \nameMethod (Ours)  & \mycmark & \mycmark & \mycmark & \textbf{2.65} & \textbf{1.21} & \textbf{2.42} & \textbf{15.64} & \textbf{10.80} & \textbf{14.90} & \textbf{56.16} & \textbf{90.42} & \textbf{56.51} \\ \bottomrule
    \end{tabular}
        }
        \vspace{-0.7 em}
        \caption{
                    \textbf{Evaluation on shape reconstruction.} 
                    We evaluate on all \nameDatasetGT sequences, using standard metrics. 
                    Columns denote the human (H), object (O), or scene (S). 
                    The ``Setup'' columns indicate whether each model (row) estimates shape for the human, object, or scene. 
        }
        \label{tab:geo_comparison}
    \end{minipage}
    \vspace{-0.5 em}
    \hfill
    \begin{minipage}{0.3\textwidth}
        \centering
        \resizebox{\textwidth}{!}{
        \begin{tabular}{l|c c c}
        \toprule
        Method & PSNR $\uparrow$ & SSIM $\uparrow$ & LPIPS $\downarrow$ \\
        \midrule
        \HSR~\cite{xue2024hsr} & 22.65 & 0.791 & 0.246 \\
        \HOLD~\cite{fan2024hold} & 17.92 & 0.646 & 0.513 \\
        \midrule
        \nameMethod (Ours) & \textbf{25.80} & \textbf{0.832} & \textbf{0.212} \\
        \bottomrule
        \end{tabular}
        }
        \vspace{-0.5 em}
        \caption{
                \textbf{Evaluation on novel-view synthesis.} 
                We evaluate on all \nameDatasetGT sequences.
                Our method (\nameMethod) provides substantially better view synthesis quality, outperforming both baselines across all metrics.
        }
         \label{tab:nvs}
        \end{minipage}
    \vspace{+0.5 em}
\end{table*}

    \zheading{Contact Estimation} %
    We leverage the recent image-based \InteractVLM~\cite{dwivedi_interactvlm_2025} model to estimate 3D body contact points from each frame.
    However, this sometimes yields false positives and temporally-inconsistent detections. 
    We tackle this by exploiting object motion. 
    Since objects move only under manipulation, any frame containing object motion is labeled as a ``contact frame,'' 
    suppressing false-positives on frames where no interaction occurs. 
    Moreover, we apply a temporal filter on raw contact predictions to improve their temporal consistency; for details see \supmat

    Then, we use the object's neural \SDF of \cref{eq:sdf_object} to 
    define a differentiable term %
    $\xi^O_{x_c} = f_{sdf}^{O}(x_c)$ 
    that attracts the body-contact points, $x_c$, onto the object. 
    We also apply contact and collision losses for physical plausibility as in \cite{tripathi2023ipman}:
    \begin{align}
        \small
        \mathcal{L}_{\text{contact}} &= \alpha_1 \tanh \left( \xi^O_{x_c} / \alpha_2 \right)^2    \quad \text{if } \xi^O_{x_c} \ge 0,      \label{eq:loss_contact}      \text{and}  \\ 
        \mathcal{L}_{\text{collision}} &= \beta_1 \tanh \left( \xi^O_{x_c} / \beta_2 \right)^2     \quad~\text{if } \xi^O_{x_c} < 0.        \label{eq:loss_collision}
    \end{align}

    \zheading{Pose Refinement via Contact}
    We use the physical losses of \cref{eq:loss_contact} and \cref{eq:loss_collision} in an optimization-based framework. 
    However, applying these from the start harms object shape due to inter-penetrations. 
    To tackle this, we take a two-stage approach. 
    In the first stage, we optimize everything using the losses of~\cref{sec:method-refinement-SDFs} (\RGB, mask, and shape cues); the physical losses are omitted. %
    In the second stage, we freeze the 
    shape networks \mbox{(\cref{eq:sdf_human} -- \cref{eq:sdf_scene})} and 
    appearance networks \mbox{(\cref{eq:rgb_human} -- \cref{eq:rgb_scene})} 
    and optimize only the human and object poses; all losses are used, including physical losses. 
    We alternate between these two stages throughout the optimization; in this way the physical losses refine poses without corrupting the object's shape.

\section{Evaluation}

First, we discuss the dataset and metrics. %
Next, we evaluate 
\nameMethod 
against 
\sota 
methods on shape reconstruction and novel-view synthesis.
Last, we 
ablate our design choices. 

\subsection{Dataset -- \nameDatasetGT}
\label{sec:dataset}
Most \HOI datasets have been captured with static cameras \cite{zhao2024imhoi, zhang2023neuraldome, bhatnagar2022behave, zhang2024hoi, huang2024intercap}. 
No existing dataset captures full-body and object interactions with a moving camera, 
similarly to cameras of
Internet videos. 
We fill this gap by capturing \nameDatasetGT, a new benchmark dataset captured with a hand-held camera moving within a volumetric 4D capture studio. 
This lets us capture monocular, moving-viewpoint \RGB video, with frames paired with 3D ground truth (GT).

\begin{figure*}
    \centering
    \vspace{-1.5 em}
    \captionsetup{type=figure}%
        \includegraphics[width=0.85 \textwidth]{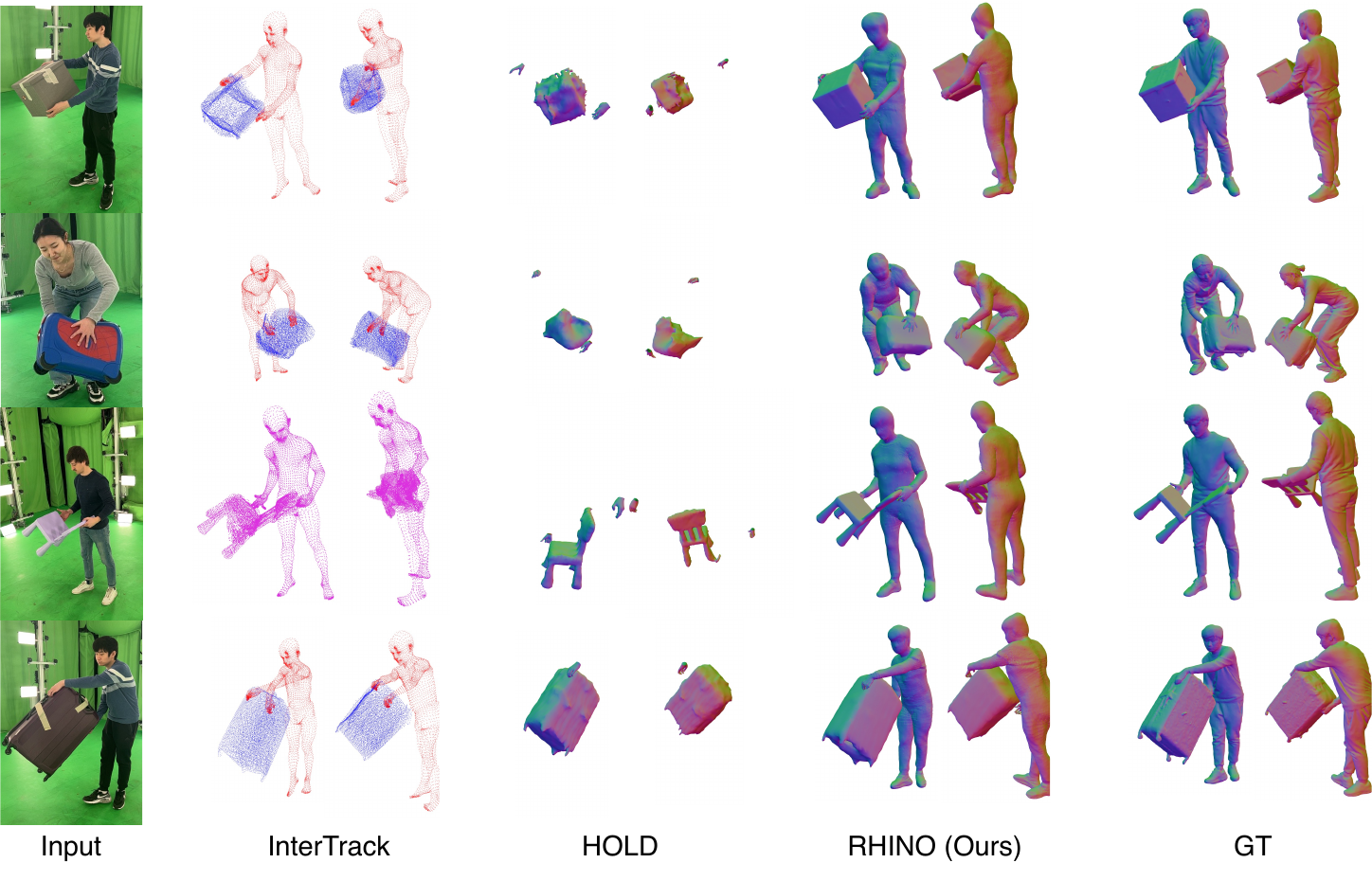}
        \vspace{-1.0 em}
    \caption{
    \textbf{Evaluation on shape reconstruction (\cref{subsec:task1-shape})} on our \nameDatasetGT dataset (\cref{sec:dataset}). 
        \HOLD~\cite{fan2024hold} struggles with noisy object poses (rows 2, 4) and fails to model 
        interaction.
        \InterTrack~\cite{xie2025intertrack} recovers reasonable object shape but fails to model the interaction due to large human and object pose errors.
        Our method (\nameMethod) faithfully recovers interactions, which lie closer to the ground truth (GT).
    }
    \label{fig:geo}
\end{figure*}

\zheading{\nameDatasetGT: Setup \& Statistics}
We use a capture studio comprising 106 synchronized cameras (53 \RGB and 53 IR cameras) at 12 MP resolution and 30 FPS. 
We capture 7 sequences, featuring 4 subjects manipulating 6 objects.

\zheading{Shape GT}
Shape is reconstructed from the raw images via a commercial software~\cite{hqvideorecon}.
To separate this shape into a human mesh and object mesh, we first detect respective masks (paired with labels) in \RGB images via \SAM~\cite{ravi2025sam2}, and then back-project masks from all views while applying majority voting to factor-out noise in \SAM mask labels.

\zheading{Camera Motion GT}
We use an \mbox{iPhone} to capture video with a moving viewpoint. 
To recover its GT motion (sequence of 6D poses), we take two steps. 
First, we attach an \mbox{AprilTag}~\cite{apriltag} to get a rough initial pose trajectory within the capture stage. 
Then, we utilize the mask loss between the projected mesh mask and the segmentation masks obtained with \SAM~\cite{ravi2025sam2}, similar to the \mbox{MultiPly} method \cite{multiply}.

\zheading{\nameDatasetITW: In-the-wild videos}
Since our annotated dataset is captured in lab settings, we also capture videos in natural indoor environments for qualitative evaluation. 
We capture 5 sequences of 3 subjects manipulating 4 objects.

\subsection{Evaluation Protocol}
    \zheading{Baselines}
        No existing method can reconstruct detailed human-object interactions in a world frame from monocular \RGB videos. 
        We compare to the following closest related work:
        \HOLD~\cite{fan2024hold} reconstructs hand-object interactions only in the camera frame, and struggles with textureless objects and rapid pose changes. 
        \HSR~\cite{xue2024hsr} reconstructs a human and a scene in a world frame, but cannot handle dynamic human-object interactions.
        \InterTrack~\cite{xie2025intertrack} predicts sparse human and object point clouds from an \RGB video, but struggles generalizing for previously unseen objects, and cannot produce a detailed reconstruction.
    
    \zheading{Evaluation Metrics}
        Shape reconstruction is evaluated via Chamfer distance (CD), Hausdorff distance (HD), and F1 score at 2 cm. 
        Baselines perform reconstruction in the camera frame, so we perform ICP to align their results to the GT mesh before computing the metrics. 
        Novel-view synthesis is evaluated via PSNR, SSIM~\cite{wang2003multiscale}, and LPIPS~\cite{zhang2018perceptual}.

\subsection{Task 1: Shape Reconstruction}
\label{subsec:task1-shape}

\begin{figure}[t]
    \centering
    \captionsetup{type=figure}%
        \includegraphics[width=0.48 \textwidth]{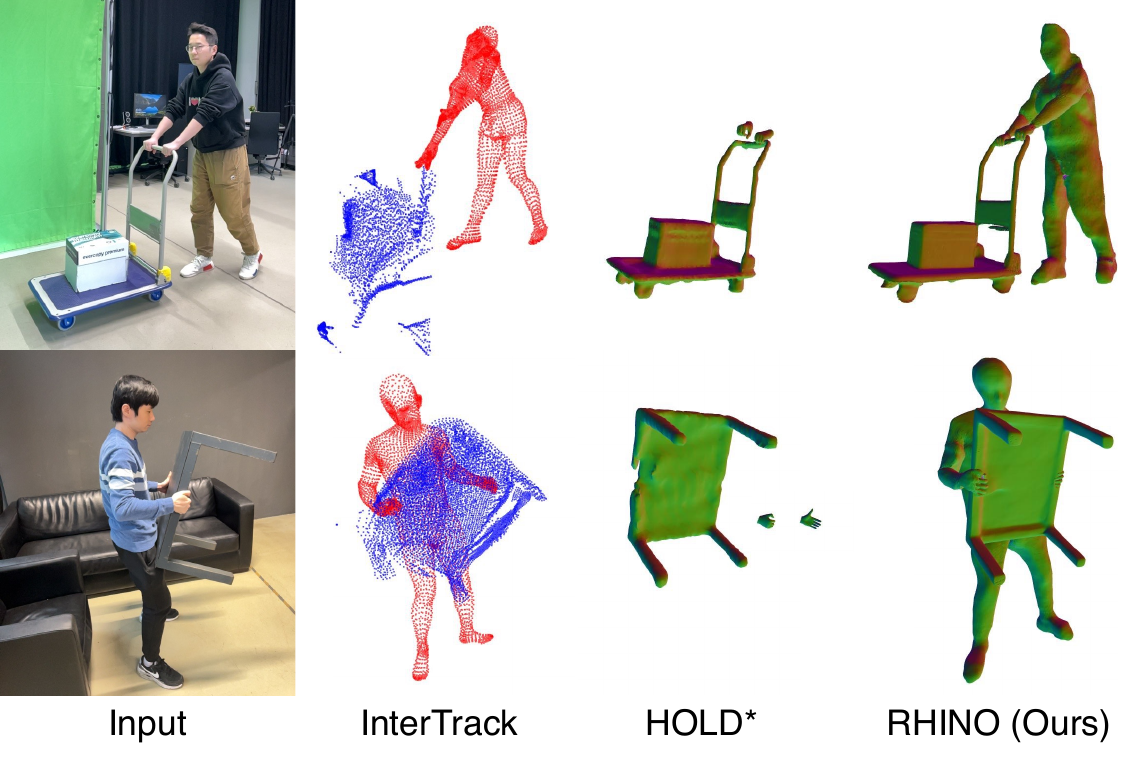}
        \vspace{-2.0 em}
    \caption{
        \textbf{Evaluation on shape reconstruction} on \nameDatasetITW. %
        \InterTrack~\cite{xie2025intertrack} fails on these out-of-distribution objects.
        While \HOLDplusRHINO (vanilla \HOLD~\cite{fan2024hold} using our method's object poses) reconstructs good object shape, it fails to model interactions.
        \nameMethod yields reasonable reconstruction reflecting the interaction.
    }
    \label{fig:geo_inthewild}
\end{figure}

    We compare to \HSR~\cite{xue2024hsr}, \HOLD~\cite{fan2024hold}, and \InterTrack~\cite{xie2025intertrack} on shape reconstruction on the \nameDatasetGT dataset. 
    As shown in ~\cref{tab:geo_comparison}, our method clearly outperforms all baselines~\cite{xie2025intertrack, fan2024hold, xue2024hsr}. %
    The difference becomes more evident in the qualitative comparisons presented in 
    \cref{fig:geo,fig:geo_inthewild}.
    \HOLD~\cite{fan2024hold} struggles reconstructing good shape when the object pose is noisy (\cref{fig:geo}, 2nd row and 4th row).
    Equipping \HOLD with accurate object pose estimation leads to significantly better object reconstruction quality in ~\cref{fig:geo_inthewild}. 
    However, it still struggles modeling hand-object interaction, 
    as the reconstructed shapes look aligned to image cues, but are misaligned \wrt each other in 3D space. 
    \InterTrack~\cite{xie2025intertrack} reconstructs reasonable shape for objects similar to its training data (\eg, chair, suitcase), 
    but struggles modeling the relative spatial configuration of the human and object. %
    For out-of-distribution objects, \InterTrack 
    struggles significantly more, 
    as shown in ~\cref{fig:geo_inthewild}.
    In contrast, our method robustly reconstructs unseen %
    objects and %
    \HOIs
    that closely resemble the ground truth.

    \begin{figure}
    \centering
        \includegraphics[width=0.49\textwidth]{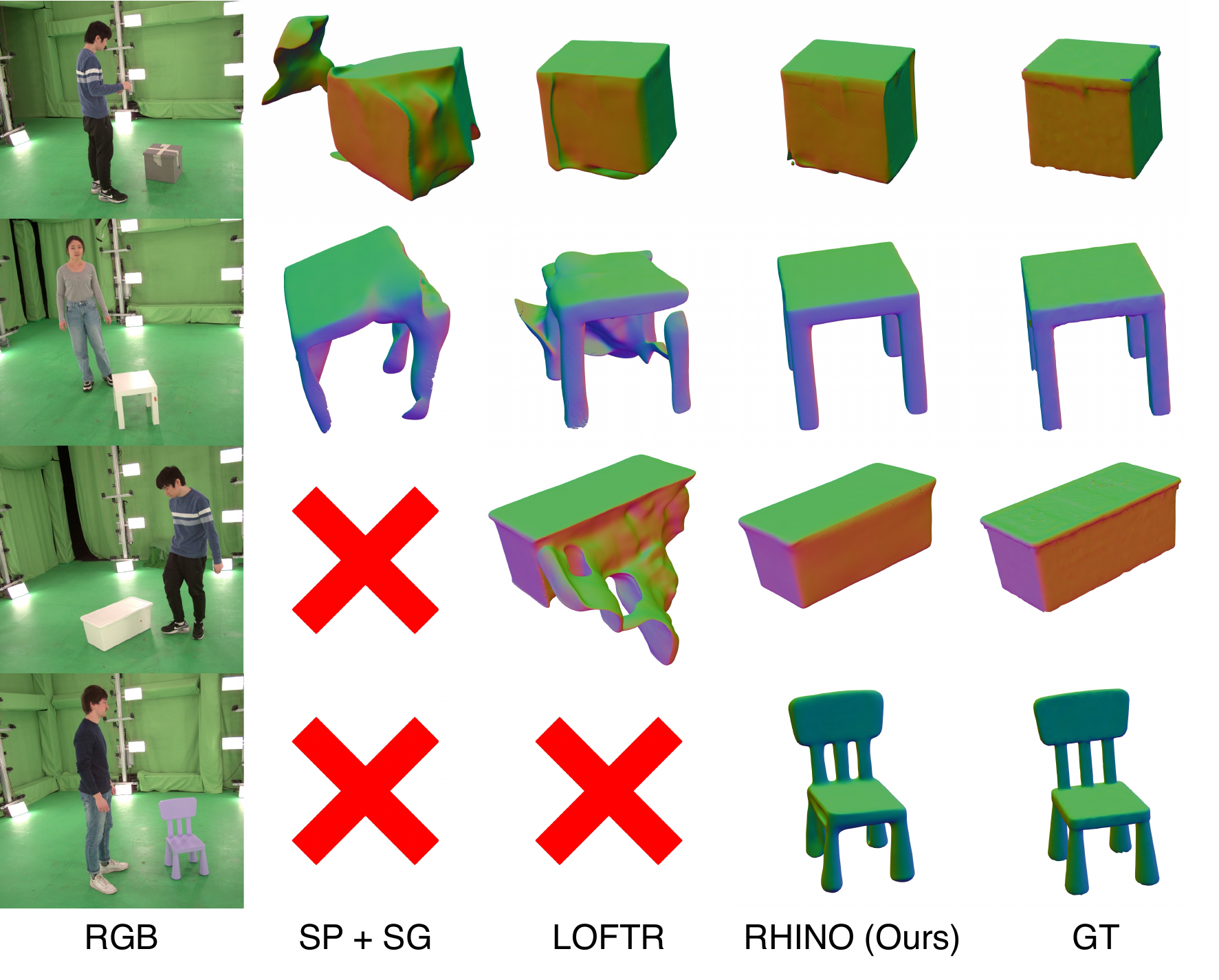}
        \vspace{-2.0 em}
    \caption{
                    \textbf{Evaluation on object pose estimation (\cref{subsec:ablations}).} %
                    We compare to {\tt SP+SG}, a \HOLD-inspired~\cite{fan2024hold} baseline that uses SuperPoints~\cite{detone18superpoint} and SuperGlue~\cite{sarlin20superglue}, and one that uses \loftr \cite{sun2021loftr}. 
                    \textcolor{red}{$\boldsymbol{\times}$}
                    denotes failed reconstruction due to wrong object poses.
    }
    \label{fig:ablation_obj_pose}
    \vspace{-1.0 em}
\end{figure}

\subsection{Task 2: Novel--View Synthesis}

We compare to \HSR~\cite{xue2024hsr} and \HOLD~\cite{fan2024hold} %
quantitatively on novel-view synthesis on \nameDatasetGT, even though this task is not our primary goal. 
\InterTrack~\cite{xie2025intertrack} is excluded from this comparison as it only outputs point clouds and does not model appearance.
As shown in ~\cref{tab:nvs}, our method produces better rendering quality than the baselines. 
For visual comparisons, please see \supmat

\subsection{Ablation Study}
\label{subsec:ablations}

    \zheading{Object Pose Estimation}
    We compare our object pose estimation, which uses MASt3R~\cite{mast3r} features, against traditional pipelines.
    These include the \SuperPoint~\cite{detone18superpoint} + \SuperGlue~\cite{sarlin20superglue} pipeline (``{\tt SP+SG}''), as used in \HOLD~\cite{fan2024hold} and OnePose~\cite{sun2022onepose}, a strong 2D feature matching baseline.
    We also compare against the dense feature \LOFTR~\cite{sun2021loftr}, used in BundleSDF~\cite{bundlesdfwen2023} and OnePose++~\cite{he2022oneposeplusplus}.
    As shown in ~\cref{fig:ablation_obj_pose} and ~\cref{tab:ablation_obj_pose}, object poses estimated with our method enable significantly more robust and accurate object reconstruction. 
    The qualitative results in ~\cref{fig:ablation_obj_pose} visually explain this performance gap.
    The SP+SG pipeline
    tends to fail for large textureless objects (\cref{fig:ablation_obj_pose}, 3rd row) due to non-repeatable keypoint detection, leading to degenerate reconstruction. 
    Similarly, \LOFTR~\cite{sun2021loftr} is not accurate enough to find correspondences for objects under complex motions (\cref{fig:ablation_obj_pose}, 4th row). In contrast, our method proves robust in these challenging scenarios.
    More visual comparisons on feature matching are shown in the 
    \purple{video on our \href{https://lxxue.github.io/RHINO}{website}}. 

    \begin{figure}
    \centering
    \captionsetup{type=figure}
        \includegraphics[width=0.49 \textwidth]{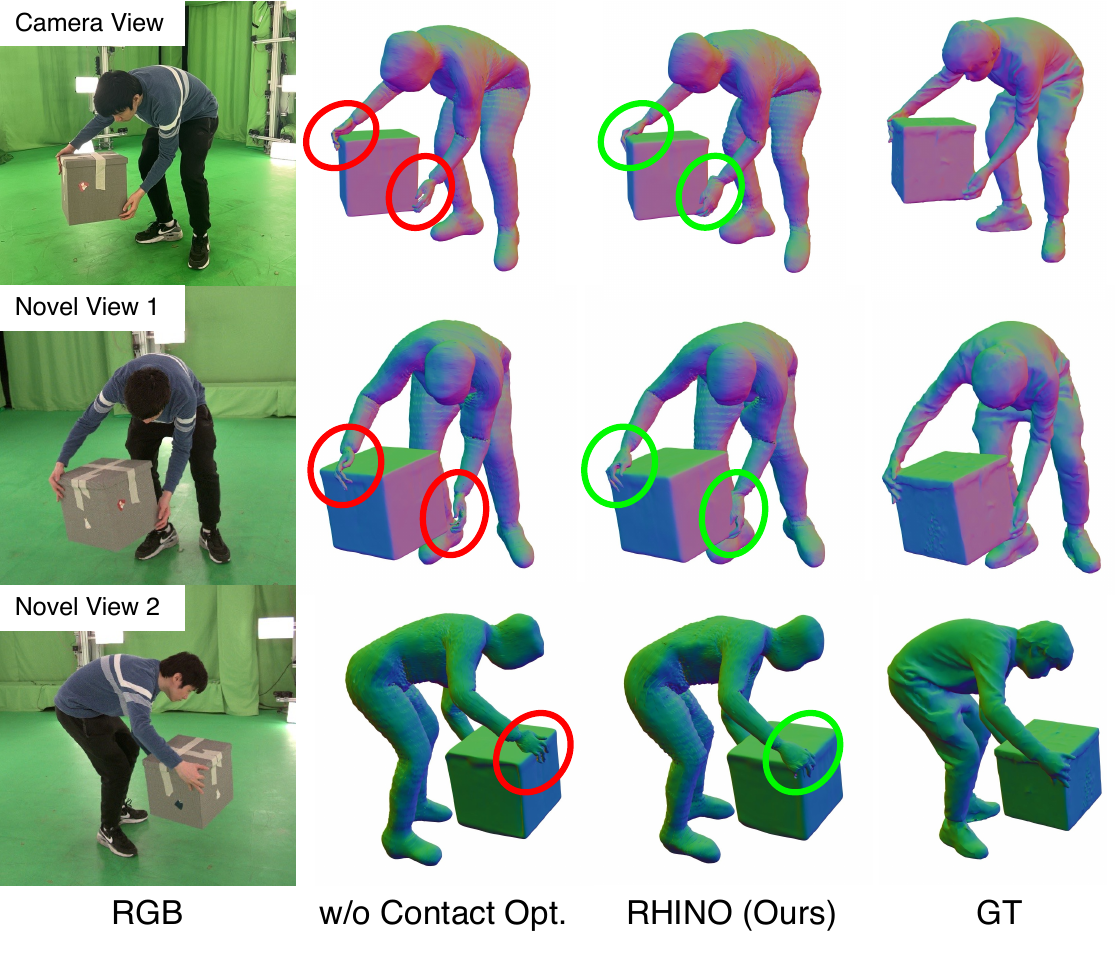}
    \caption{
        \textbf{Effects of contact.} 
        We show reconstructions of our framework with (``{\tt RHINO (Ours)}'') and 
        without (``{\tt w/o Contact Opt}'') the physical losses of \cref{eq:loss_contact} and \cref{eq:loss_collision}. 
        For a bigger version with zoom-in impressions, see \supmat 
    }
    \label{fig:ablation_contact}
    \vspace{+1.0 em}
\end{figure}

    \zheading{Motion Disentanglement (MD)}
    To evaluate %
    our camera--object motion disentanglement (\cref{sec:method-frame-alignment}), we compare against a variant that uses apparent object motion without removing the camera component. As shown in \cref{tab:ablation_motion_decomposition}, removing MD (``{\tt w/o MD}'') drastically worsens all metrics, showing that MD is crucial for world-frame reconstruction.

    \zheading{Contact Refinement}
    To investigate the importance of our contact-based pose refinement, we compare our full model (``{\tt RHINO (Ours)}") to a version without this step (``{\tt w/o Contact Opt}"), and show results in \cref{fig:ablation_contact}.
    As discussed in \cref{sec:method-contact}, while initial reconstructions may align well with pixels (\cref{fig:ablation_contact}, 1st row), they often suffer physical implausibility due to depth ambiguity and occlusions, causing hands to hover over or penetrate an object.
    \Cref{fig:ablation_contact} clearly shows these artifacts in novel views (red circles of the 2nd and 3rd row).
    On the other hand, our full RHINO model leverages contact priors and produces reconstructions in which the hands establish firm, realistic contact with the object, closely matching the ground truth.
    We quantitatively evaluate %
    in \cref{tab:ablation_contact} by reporting penetration depth (PD), contact precision, recall, and F1. Contact refinement halves the PD and nearly triples the recall, which shows its importance for physical plausibility.
    
    \begin{table}[t]
    \centering
    \vspace{-0.2 em}
    \resizebox{0.90 \columnwidth}{!}{%
        \begin{tabular}{l c c c c}
            \toprule
             & CD [cm] $\downarrow$ & HD [cm] $\downarrow$ & F1 Score [\%] $\uparrow$ \\ 
            \midrule
            SP \cite{detone18superpoint} + SG \cite{sarlin20superglue}  & 4.25 & 25.43 & 60.06 \\
            \loftr \cite{sun2021loftr}                                  & 3.97 & 20.19 & 62.80  \\
            \midrule
            \nameMethod (Ours) & \textbf{1.09}  & \textbf{10.94} & \textbf{91.38} \\
            \bottomrule
        \end{tabular}
    }
    \vspace{-0.7 em}
    \caption{
                \textbf{Evaluation on object pose estimation (\cref{subsec:ablations}).} 
                We compare to {\tt SP+SG}, a \HOLD-inspired~\cite{fan2024hold} baseline that uses \SuperPoint~\cite{detone18superpoint} and \SuperGlue~\cite{sarlin20superglue}, and one that uses \loftr \cite{sun2021loftr}. 
                Results are reported for \nameDatasetGT sequences for which all baselines 
                do not fail; 
                see the list of sequences in \supmat 
    }
    \label{tab:ablation_obj_pose}
\end{table}

    \begin{table}[t]
    \centering
    \resizebox{0.95\columnwidth}{!}{
    \begin{tabular}{l|c c c c}
        \toprule
        Method & CD [cm] $\downarrow$ & F1@2cm [\%] $\uparrow$ & PSNR $\uparrow$ & LPIPS $\downarrow$ \\
        \midrule
        w/o MD & 10.21 & 26.32 & 22.89 & 0.306 \\
        \nameMethod (Ours) & \textbf{2.65} & \textbf{56.16} & \textbf{25.80} & \textbf{0.212} \\
        \bottomrule
    \end{tabular}
    }
    \vspace{-0.7em}
    \caption{
        \textbf{Ablation on motion disentanglement (MD) (\cref{subsec:ablations}).}
        Removing motion disentanglement leads to a large drop across all metrics, confirming it is essential for world-frame reconstruction.
    }
    \label{tab:ablation_motion_decomposition}
\end{table}

    \begin{table}[t]
    \centering
    \resizebox{0.95\columnwidth}{!}{
    \begin{tabular}{l|c c c c}
        \toprule
        Method & PD [cm] $\downarrow$ & Precision [\%] $\uparrow$ & Recall [\%] $\uparrow$ & F1 [\%] $\uparrow$ \\
        \midrule
        w/o Contact Opt. & 1.088 & 24.06 & 18.39 & 20.84 \\
        \nameMethod (Ours) & \textbf{0.477} & \textbf{25.67} & \textbf{63.57} & \textbf{36.57} \\
        \bottomrule
    \end{tabular}
    }
    \vspace{-0.7em}
    \caption{
        \textbf{Ablation on contact refinement (\cref{subsec:ablations}).}
        Refining pose via contact 
        reduces penetration depth (PD) and increases contact recall and F1, showing its importance for physical plausibility.
    }
    \label{tab:ablation_contact}
    \vspace{-1.0 em}
\end{table}

\section{Conclusion}

In this paper, 
we present \nameMethod, 
a novel framework that reconstructs %
4D human-object interaction scenes from a single monocular video with a moving viewpoint.
First, we develop a robust 3D-aware methodology that estimates an initial object and scene shape, human motion, and disentangled object and camera motion.
Then, we refine these using a compositional neural field with per-component \SDFs. 
This not only captures shape details, but also enables defining a contact loss with differentiable \SDF-based distances, ensuring physically-plausible interactions. 
For evaluation, we capture \nameDatasetGT, a new video dataset of human-object interactions with 4D ground truth. 
\nameMethod is able to reconstruct high-quality human-object interactions in challenging scenarios such as occlusions and complex motions.

\smallskip

\qheading{Future Work}
Our current framework is designed for interactions involving a single person and a single rigid object. 
Extending it to handle more complex scenes with multiple interacting humans or objects, %
remains a significant challenge. 
Moreover, our framework assumes object rigidity. 
A valuable direction for future work would be to incorporate the reconstruction of non-rigid objects, such as articulated objects, to capture a wider range of real-world interactions.
Finally, our current optimization process is slow. 
Speeding it up to enable fast 4D capture would be crucial for applications in AR/VR and robotics. We also see potential for improving robustness to sparse observations, as our method currently performs best with good object coverage.

\section*{Acknowledgements}
We thank all participants of the captured dataset.
We also thank Otmar Hilliges for initiating and supporting this work.
The ETH part of the team
was partly supported by the Swiss SERI Consolidation Grant AI-PERCEIVE.
Manuel Kaufmann was partly supported by the ETH AI center.
Compute was partly performed on the ETH Zurich Euler cluster.
The UvA part of the team was supported by the ERC Starting Grant (project \mbox{STRIPES}, \mbox{101165317}, PI:~D.~Tzionas), by a research gift from Google, by the NVIDIA Academic Grant Program,
and by compute offered
by the \mbox{EuroHPC} JU
supercomputers \mbox{LEONARDO} (project ID: \mbox{EHPC-AI-2024A06-077})
and \mbox{JUPITER} (project ID: \mbox{e-reg-2025r02-393}),
and by the Dutch
e-infrastructure
SURF
(project IDs: \mbox{EINF-10235} and \mbox{EINF-16839}).

{
    \small
    \bibliographystyle{config/ieeenat_fullname}
    \bibliography{config/BIB}

@String(PAMI 	=	{{Transactions on Pattern Analysis and Machine Intelligence (TPAMI)}})

@String(IJCV 	=	{{International Journal of Computer Vision (IJCV)}})

@String(CVPR 	=	{{Computer Vision and Pattern Recognition (CVPR)}})

@String(CVPRW 	=	{{Computer Vision and Pattern Recognition Workshops (CVPRw)}})

@String(ICCV 	=	{{International Conference on Computer Vision ({ICCV})}})

@String(ECCV 	=	{{European Conference on Computer Vision (ECCV)}})

@String(BMVC 	=	{{British Machine Vision Conference (BMVC)}})

@String(TOG 	=	{{Transactions on Graphics (TOG)}})

@String(ICLR 	=	{{International Conference on Learning Representations (ICLR)}})

@STRING(I3DV 	=	{{International Conference on {3D} Vision (3DV)}})

@STRING(CGF 	=	{{Computer Graphics Forum (CGF)}})

@String(NEURIPS	=   {{Neural Information Processing Systems (NeurIPS)}})

@String(ICRA	= 	{{International Conference on Robotics and Automation (ICRA)}})

@String(ICML    = {{International Conference on Machine Learning (ICML)}})

@article{hqvideorecon,
author = {Collet, Alvaro and Chuang, Ming and Sweeney, Pat and Gillett, Don and Evseev, Dennis and Calabrese, David and Hoppe, Hugues and Kirk, Adam and Sullivan, Steve},
title = {High-quality streamable free-viewpoint video},
year = {2015},
issue_date = {August 2015},
publisher = {Association for Computing Machinery},
address = {New York, NY, USA},
volume = {34},
number = {4},
issn = {0730-0301},
url = {https://doi.org/10.1145/2766945},
doi = {10.1145/2766945},
journal = TOG,
month = jul,
articleno = {69},
numpages = {13}
}

@misc{ren2024grounding,
      title={Grounding {DINO 1.5}: {A}dvance the ``Edge" of Open-Set Object Detection}, 
      author={Tianhe Ren and Qing Jiang and Shilong Liu and Zhaoyang Zeng and Wenlong Liu and Han Gao and Hongjie Huang and Zhengyu Ma and Xiaoke Jiang and Yihao Chen and Yuda Xiong and Hao Zhang and Feng Li and Peijun Tang and Kent Yu and Lei Zhanget},
      year={2024},
      howpublished={{arXiv}:2405.10300},
}

@INPROCEEDINGS{apriltag,
  author={Olson, Edwin},
  booktitle=ICRA, 
  title={{AprilTag}: {A} robust and flexible visual fiducial system}, 
  year={2011},
  volume={},
  number={},
}

@inproceedings{tripathi2023ipman,
    title = {{3D} Human Pose Estimation via Intuitive Physics},
    author = {Tripathi, Shashank and M{\"u}ller, Lea and Huang, Chun-Hao P. and Taheri Omid and Black, Michael J. and Tzionas, Dimitrios},
    booktitle = CVPR,
    pages = {4713--4725},
    year = {2023},
    url = {https://ipman.is.tue.mpg.de}
}

@inproceedings{dwivedi_interactvlm_2025,
  title     = {{InteractVLM}: {3D} Interaction Reasoning from {2D} Foundational Models},
  author    = {Dwivedi, Sai Kumar and Antić, Dimitrije and Tripathi, Shashank and Taheri, Omid and Schmid, Cordelia and Black, Michael J. and Tzionas, Dimitrios},
  booktitle = CVPR,
  year      = {2025},
}

@InProceedings{bundlesdfwen2023,
    author      =   {Bowen Wen and Jonathan Tremblay and Valts Blukis and Stephen Tyree and Thomas M\"{u}ller and Alex Evans and Dieter Fox and Jan Kautz and Stan Birchfield},
    title       =   {{BundleSDF}: {N}eural 6-{DoF} Tracking and {3D} Reconstruction of Unknown Objects},
    booktitle   =   CVPR,
    year        =   {2023},
}

@InProceedings{sun2022onepose,
    title       =   {{OnePose}: {O}ne-Shot Object Pose Estimation without {CAD} Models},
    author      =   {Sun, Jiaming and Wang, Zihao and Zhang, Siyu and He, Xingyi and Zhao, Hongcheng and Zhang, Guofeng and Zhou, Xiaowei},
    booktitle   =   CVPR,
    year        =   {2022},
}

@inproceedings{antic2025sdfit,
    title     = {{SDFit}: {3D} Object Pose and Shape by Fitting a Morphable {SDF} to a Single Image},
    author    = {Anti\'{c}, Dimitrije and Paschalidis, Georgios and Tripathi, Shashank and Gevers, Theo and Dwivedi, Sai Kumar and Tzionas, Dimitrios},
    booktitle = ICCV,
    year      = {2025},
}

@InProceedings{
    he2022oneposeplusplus,
    title={{OnePose++}: {K}eypoint-Free One-Shot Object Pose Estimation without {CAD} Models},
    author={Xingyi He and Jiaming Sun and Yuang Wang and Di Huang and Hujun Bao and Xiaowei Zhou},
    booktitle=NeurIPS,
    year={2022}
}

@InProceedings{mast3r,
      title={Grounding Image Matching in {3D} with {MASt3R}}, 
      author={Vincent Leroy and Yohann Cabon and Jerome Revaud},
      booktitle = ECCV,
      year = {2024}
}

@InProceedings{teed2021droid,
  title={{DROID-SLAM}: {D}eep Visual {SLAM} for Monocular, Stereo, and {RGB-D} Cameras},
  author={Teed, Zachary and Deng, Jia},
  booktitle=NEURIPS,
  year={2021}
}

@InProceedings{dust3r,
      title={{DUSt3R}: {G}eometric {3D} Vision Made Easy}, 
      author={Shuzhe Wang and Vincent Leroy and Yohann Cabon and Boris Chidlovskii and Jerome Revaud},
      booktitle = CVPR,
      year = {2024}
}

@InProceedings{wang2025vggt,
  title={{VGGT}: {V}isual {G}eometry {G}rounded {T}ransformer},
  author={Wang, Jianyuan and Chen, Minghao and Karaev, Nikita and Vedaldi, Andrea and Rupprecht, Christian and Novotny, David},
  booktitle=CVPR,
  year={2025}
}

@InProceedings{chen2025back,
  title={Back on Track: {B}undle Adjustment for Dynamic Scene Reconstruction},
  author={Chen, Weirong and Zhang, Ganlin and Wimbauer, Felix and Wang, Rui and Araslanov, Nikita and Vedaldi, Andrea and Cremers, Daniel},
  booktitle=ICCV,
  year={2025}
}

@InProceedings{qian20233dgsavatar,
   title={{3DGS-Avatar}: {A}nimatable Avatars via Deformable {3D} Gaussian Splatting}, 
   author={Zhiyin Qian and Shaofei Wang and Marko Mihajlovic and Andreas Geiger and Siyu Tang},
   booktitle=CVPR, 
   year={2024},
}

@InProceedings{guo2025vid2avatarpro,
      title     = {{Vid2Avatar-Pro}: {A}uthentic Avatar from Videos in the Wild via Universal Prior},
      author    = {Guo, Chen and Li, Junxuan and Kant, Yash and Sheikh, Yaser and Saito, Shunsuke and Cao, Chen},
      booktitle = CVPR,
      year      = {2025},
}

@InProceedings{fan2024hold,
  title={{HOLD}: {C}ategory-agnostic {3D} reconstruction of interacting hands and objects from video},
  author={Fan, Zicong and Parelli, Maria and Kadoglou, Maria Eleni and Kocabas, Muhammed and Chen, Xu and Black, Michael J and Hilliges, Otmar},
  booktitle=CVPR,
  year={2024}
}

@InProceedings{xue2024hsr,
    author      =   {Xue, Lixin and Guo, Chen and Zheng, Chengwei and Wang, Fangjinhua and Jiang, Tianjian and Ho, Hsuan-I and Kaufmann, Manuel and Song, Jie and Hilliges Otmar},
    title       =   {{HSR}: {H}olistic {3D} Human-Scene Reconstruction from Monocular Videos},
    booktitle   =   ECCV,
    year        =   {2024}
}

@InProceedings{pan2024glomap,
    author={Pan, Linfei and Baráth, Dániel and Pollefeys, Marc and Sch\"{o}nberger, Johannes Lutz},
    title={Global Structure-from-Motion Revisited},
    booktitle=ECCV,
    year={2024},
}

@InProceedings{detone18superpoint,
  author    = {Daniel DeTone and
               Tomasz Malisiewicz and
               Andrew Rabinovich},
  title     = {{SuperPoint}: {S}elf-Supervised Interest Point Detection and Description},
  booktitle = CVPRW,
  year      = {2018}
}

@InProceedings{sun2021loftr,
  title={{LoFTR}: {D}etector-Free Local Feature Matching with Transformers},
  author={Sun, Jiaming and Shen, Zehong and Wang, Yuang and Bao, Hujun and Zhou, Xiaowei},
  booktitle=CVPR,
  year={2021}
}

@article{umeyama1991estimation,
  author={Umeyama, Shinji},
  journal=PAMI, 
  title={Least-squares estimation of transformation parameters between two point patterns}, 
  year={1991}
}

@InProceedings{guo2023vid2avatar,
      title={{Vid2Avatar}: {3D} Avatar Reconstruction from Videos in the Wild via Self-supervised Scene Decomposition},
      author={Guo, Chen and Jiang, Tianjian and Chen, Xu and Song, Jie and Hilliges, Otmar}, 
      booktitle = CVPR,
      year      = {2023},
}

@InProceedings{sun2024aios,
    author       = {Qingping Sun and
                    Yanjun Wang and
                    Ailing Zeng and
                    Wanqi Yin and
                    Chen Wei and
                    Wenjia Wang and
                    Haiyi Mei and
                    Chi{-}Sing Leung and
                    Ziwei Liu and
                    Lei Yang and
                    Zhongang Cai},
    title        = {{AiOS}: {A}ll-in-One-Stage Expressive Human Pose and Shape Estimation},
    booktitle    = CVPR,
    year         = {2024}
}

@InProceedings{khirodkar2024sapiens,
    author       = {Rawal Khirodkar and
                    Timur M. Bagautdinov and
                    Julieta Martinez and
                    Su Zhaoen and
                    Austin James and
                    Peter Selednik and
                    Stuart Anderson and
                    Shunsuke Saito},
    title        = {Sapiens: {F}oundation for Human Vision Models},
    booktitle    = ECCV,
    year         = {2024}
}

@InProceedings{jiang2022neuman,
    author       = {Wei Jiang and
                    Kwang Moo Yi and
                    Golnoosh Samei and
                    Oncel Tuzel and
                    Anurag Ranjan},
    title        = {{NeuMan}: {N}eural Human Radiance Field from a Single Video},
    booktitle    = ECCV,
    year         = {2022}
}

@InProceedings{hassan2019resolving,
  title={Resolving {3D} human pose ambiguities with {3D} scene constraints},
  author={Hassan, Mohamed and Choutas, Vasileios and Tzionas, Dimitrios and Black, Michael J},
  booktitle=ICCV,
  year={2019}
}

@InProceedings{huang2022capturing,
  title={Capturing and inferring dense full-body human-scene contact},
  author={Huang, Chun-Hao P and Yi, Hongwei and H{\"o}schle, Markus and Safroshkin, Matvey and Alexiadis, Tsvetelina and Polikovsky, Senya and Scharstein, Daniel and Black, Michael J},
  booktitle=CVPR,
  year={2022}
}

@InProceedings{dai2022hsc4d,
  title={{HSC4D}: {H}uman-centered {4D} Scene Capture in Large-scale Indoor-outdoor Space Using Wearable IMUs and LiDAR},
  author={Dai, Yudi and Lin, Yitai and Wen, Chenglu and Shen, Siqi and Xu, Lan and Yu, Jingyi and Ma, Yuexin and Wang, Cheng},
  booktitle=CVPR,
  year={2022}
}

@InProceedings{guzov2021human,
  title={Human {POSE}itioning System {(HPS)}: {3D} Human Pose Estimation and Self-Localization in Large Scenes From Body-Mounted Sensors},
  author={Guzov, Vladimir and Mir, Aymen and Sattler, Torsten and Pons-Moll, Gerard},
  booktitle=CVPR,
  year={2021}
}

@InProceedings{kaufmann2023emdb,
  title={{EMDB}: {T}he Electromagnetic Database of Global {3D} Human Pose and Shape in the Wild},
  author={Kaufmann, Manuel and Song, Jie and Guo, Chen and Shen, Kaiyue and Jiang, Tianjian and Tang, Chengcheng and Z{\'a}rate, Juan Jos{\'e} and Hilliges, Otmar},
  booktitle=ICCV,
  year={2023}
}

@InProceedings{zhang2022egobody,
  title={{EgoBody}: {H}uman Body Shape and Motion of Interacting People from Head-Mounted Devices},
  author={Zhang, Siwei and Ma, Qianli and Zhang, Yan and Qian, Zhiyin and Kwon, Taein and Pollefeys, Marc and Bogo, Federica and Tang, Siyu},
  booktitle=ECCV,
  year={2022},
}

@InProceedings{song2023total,
  title={{Total-Recon}: {D}eformable Scene Reconstruction for Embodied View Synthesis},
  author={Song, Chonghyuk and Yang, Gengshan and Deng, Kangle and Zhu, Jun-Yan and Ramanan, Deva},
  booktitle=ICCV,
  year={2023}
}

@InProceedings{yang2023ppr,
  title={{PPR}: {P}hysically Plausible Reconstruction from Monocular Videos},
  author={Yang, Gengshan and Yang, Shuo and Zhang, John Z and Manchester, Zachary and Ramanan, Deva},
  booktitle=ICCV,
  year={2023}
}

@InProceedings{bhatnagar2022behave,
  title={{BEHAVE}: {D}ataset and Method for Tracking Human Object Interactions},
  author={Bhatnagar, Bharat Lal and Xie, Xianghui and Petrov, Ilya A and Sminchisescu, Cristian and Theobalt, Christian and Pons-Moll, Gerard},
  booktitle=CVPR,
  year={2022}
}

@InProceedings{nam2024joint,
  title={Joint Reconstruction of {3D} Human and Object via Contact-Based Refinement Transformer},
  author={Nam, Hyeongjin and Jung, Daniel Sungho and Moon, Gyeongsik and Lee, Kyoung Mu},
  booktitle=CVPR,
  year={2024}
}

@InProceedings{xie2022chore,
  title={{CHORE}: {C}ontact, Human and Object Reconstruction from a Single {RGB} Image},
  author={Xie, Xianghui and Bhatnagar, Bharat Lal and Pons-Moll, Gerard},
  booktitle=ECCV,
  year={2022},
}

@InProceedings{xie2024template,
  title={Template free reconstruction of human-object interaction with procedural interaction generation},
  author={Xie, Xianghui and Bhatnagar, Bharat Lal and Lenssen, Jan Eric and Pons-Moll, Gerard},
  booktitle=CVPR,
  year={2024}
}

@InProceedings{wen2024foundationpose,
  title={{FoundationPose}: {U}nified {6D} Pose Estimation and Tracking of Novel Objects},
  author={Wen, Bowen and Yang, Wei and Kautz, Jan and Birchfield, Stan},
  booktitle=CVPR,
  year={2024}
}

@InProceedings{xie2025intertrack,
  title={{InterTrack}: {T}racking Human Object Interaction without Object Templates},
  author={Xianghui Xie and Jan Eric Lenssen and Gerard Pons-Moll},
  booktitle=I3DV,
  year={2025}
}

@InProceedings{multiply,
      title={{MultiPly}: {R}econstruction of Multiple People from Monocular Video in the Wild}, 
      author={Jiang, Zeren and Guo, Chen and Kaufmann, Manuel and Jiang, Tianjian and Valentin, Julien and Hilliges, Otmar and Song, Jie},
      booktitle = CVPR,
      year      = {2024}
}

@InProceedings{ravi2025sam2,
    title={{SAM 2}: {S}egment Anything in Images and Videos},
    author={Ravi, Nikhila and Gabeur, Valentin and Hu, Yuan-Ting and Hu, Ronghang and Ryali, Chaitanya and Ma, Tengyu and Khedr, Haitham and R{\"a}dle, Roman and Rolland, Chloe and Gustafson, Laura and Mintun, Eric and Pan, Junting and Alwala, Kalyan Vasudev and Carion, Nicolas and Wu, Chao-Yuan and Girshick et.~el., Ross},
    booktitle=ICLR,
    year={2025}
}

@InProceedings{muller2025reconstructing,
  title={Reconstructing People, Places, and Cameras},
  author={M{\"u}ller, Lea and Choi, Hongsuk and Zhang, Anthony and Yi, Brent and Malik, Jitendra and Kanazawa, Angjoo},
  booktitle = CVPR,
  year={2025}
}

@InProceedings{gopal2025betsu,
  title={{Betsu-Betsu}: {M}ulti-View Separable {3D} Reconstruction of Two Interacting Objects},
  author={Gopal, Suhas and Dabral, Rishabh and Golyanik, Vladislav and Theobalt, Christian},
  booktitle=I3DV,
  year={2025}
}

@inproceedings{caraffa2024freeze, 
  title={{FreeZe}: {T}raining-free zero-shot {6D} pose estimation with geometric and vision foundation models}, 
  author = {Caraffa, Andrea and Boscaini, Davide and Hamza, Amir and Poiesi, Fabio}, 
  booktitle = ECCV, 
  year = {2024} 
}

@inproceedings{moon2024genflow,
  title={{GenFlow}: {G}eneralizable Recurrent Flow for {6D} Pose Refinement of Novel Objects},
  author={Moon, Sungphill and Son, Hyeontae and Hur, Dongcheol and Kim, Sangwook},
  booktitle=CVPR,
  year={2024}
}

@InProceedings{Di_2021_ICCV,
author = {Di, Yan and Manhardt, Fabian and Wang, Gu and Ji, Xiangyang and Navab, Nassir and Tombari, Federico},
title = {{SO-Pose}: {E}xploiting Self-Occlusion for Direct {6D} Pose Estimation},
booktitle = ICCV,
year = {2021},
}

@inproceedings{lee2025any6d,
    title     = {{Any6D}: {M}odel-free {6D} Pose Estimation of Novel Objects},
    author    = {Lee, Taeyeop and Wen, Bowen and Kang, Minjun and Kang, Gyuree and Kweon, In So and Yoon, Kuk-Jin},
    booktitle = CVPR,
    year      = {2025},
}

@InProceedings{Lin_2024_CVPR,
    author    = {Lin, Jiehong and Liu, Lihua and Lu, Dekun and Jia, Kui},
    title     = {{SAM-6D}: {S}egment Anything Model Meets Zero-Shot {6D} Object Pose Estimation},
    booktitle = CVPR,
    year      = {2024},
}

@inproceedings{wang2019normalized,
  title={Normalized object coordinate space for category-level {6D} object pose and size estimation},
  author={Wang, He and Sridhar, Srinath and Huang, Jingwei and Valentin, Julien and Song, Shuran and Guibas, Leonidas J.},
  booktitle=CVPR,
  year={2019}
}

@inproceedings{sarlin20superglue,
  author    = {Paul-Edouard Sarlin and
               Daniel DeTone and
               Tomasz Malisiewicz and
               Andrew Rabinovich},
  title     = {{SuperGlue}: {L}earning Feature Matching with Graph Neural Networks},
  booktitle = CVPR,
  year      = {2020},
}

@article{huang2024intercap, 
    title        = {{InterCap}: {J}oint Markerless {3D} Tracking of Humans and Objects in Interaction from Multi-view {RGB-D} Images}, 
    author       = {Huang, Yinghao and Taheri, Omid and Black, Michael J. and Tzionas, Dimitrios}, 
    journal      = IJCV, 
    year         = {2024}
}

@inproceedings{weng2021holistic,
  title={Holistic {3D} human and scene mesh estimation from single view images},
  author={Weng, Zhenzhen and Yeung, Serena},
  booktitle=CVPR,
  year={2021}
}

@inproceedings{guo2020single,
  title={Single path one-shot neural architecture search with uniform sampling},
  author={Guo, Zichao and Zhang, Xiangyu and Mu, Haoyuan and Heng, Wen and Liu, Zechun and Wei, Yichen and Sun, Jian},
  booktitle=ECCV,
  year={2020},
}

@inproceedings{jiang2022neuralhofusion,
  title={{NeuralHOFusion}: {N}eural Volumetric Rendering under Human-object Interactions},
  author={Jiang, Yuheng and Jiang, Suyi and Sun, Guoxing and Su, Zhuo and Guo, Kaiwen and Wu, Minye and Yu, Jingyi and Xu, Lan},
  booktitle=CVPR,
  year={2022}
}

@inproceedings{zhang2023neuraldome,
  title={{NeuralDome}: {A} Neural Modeling Pipeline on Multi-View Human-Object Interactions},
  author={Zhang, Juze and Luo, Haimin and Yang, Hongdi and Xu, Xinru and Wu, Qianyang and Shi, Ye and Yu, Jingyi and Xu, Lan and Wang, Jingya},
  booktitle=CVPR,
  year={2023}
}

@inproceedings{zhao2024imhoi,
  title={{I'M HOI}: {I}nertia-Aware Monocular Capture of {3D} Human-Object Interactions},
  author={Zhao, Chengfeng and Zhang, Juze and Du, Jiashen and Shan, Ziwei and Wang, Junye and Yu, Jingyi and Wang, Jingya and Xu, Lan},
  booktitle=CVPR,
  year={2024}
}

@inproceedings{
      zhang2024hoi,
      title={{HOI-M3}: {C}apture Multiple Humans and Objects Interaction within Contextual Environment},
      author={Zhang, Juze and Zhang, Jingyan and Song, Zining and Shi, Zhanhe and Zhao, Chengfeng and Shi, Ye and Yu, Jingyi and Xu, Lan and Wang, Jingya},
      booktitle=CVPR,
      year={2024}
}

@misc{chen2025human3r,
    title={{Human3R}: {E}veryone Everywhere All at Once},
    author={Chen, Yue and Chen, Xingyu and Xue, Yuxuan and Chen, Anpei and Xiu, Yuliang and Gerard, Pons-Moll},
    howpublished={{arXiv}:2510.06219},
    year={2025}
    }

@inproceedings{rojas2025hamst3rhumanawaremultiviewstereo,
      title={{HAMSt3R}: {H}uman-Aware Multi-view Stereo {3D} Reconstruction}, 
      author={Sara Rojas and Matthieu Armando and Bernard Ghamen and Philippe Weinzaepfel and Vincent Leroy and Gregory Rogez},
      year={2025},
      booktitle=ICCV,
}

@misc{liu2025joint,
    title={Joint Optimization for {4D} Human-Scene Reconstruction in the Wild},
    author={Liu, Zhizheng and Lin, Joe and Wu, Wayne and Zhou, Bolei},
    howpublished={{arXiv}:2501.02158},
    year={2025}
}

@inproceedings{SceneAware_EG2023,
  title={Scene-Aware {3D} Multi-Human Motion Capture from a Single Camera}, 
  author={Diogo C. Luvizon and Marc Habermann and Vladislav Golyanik and Adam Kortylewski and Christian Theobalt},
  booktitle=CGF,
  year={2023},
}

@inproceedings{Li_3DV2022, 
    title={{MoCapDeform}: {M}onocular {3D} Human Motion Capture in Deformable Scenes}, 
    author={Zhi Li and Soshi Shimada and Bernt Schiele and Christian Theobalt and Vladislav Golyanik}, 
    booktitle = I3DV, 
    year={2022} 
}

@inproceedings{SMPL-X:2019,
  title = {Expressive Body Capture: {3D} Hands, Face, and Body from a Single Image},
  author = {Pavlakos, Georgios and Choutas, Vasileios and Ghorbani, Nima and Bolkart, Timo and Osman, Ahmed A. A. and Tzionas, Dimitrios and Black, Michael J.},
  booktitle = CVPR,
  year = {2019}
}

@inproceedings{wang2003multiscale,
  title={Multiscale structural similarity for image quality assessment},
  author={Wang, Zhou and Simoncelli, Eero P and Bovik, Alan C},
  booktitle={Asilomar Conference on Signals, Systems \& Computers},
  volume={2},
  year={2003}
}

@inproceedings{zhang2018perceptual,
  title={The Unreasonable Effectiveness of Deep Features as a Perceptual Metric},
  author={Zhang, Richard and Isola, Phillip and Efros, Alexei A and Shechtman, Eli and Wang, Oliver},
  booktitle={CVPR},
  year={2018}
}

@InProceedings{zhang2025odhsr,
    author    = {Zhang, Zetong and Kaufmann, Manuel and Xue, Lixin and Song, Jie and Oswald, Martin R.},
    title     = {{ODHSR}: {O}nline Dense {3D} Reconstruction of Humans and Scenes from Monocular Videos},
    booktitle = CVPR,
    year      = {2025},
}

@InProceedings{icml2020_igr,
 author = {Gropp, Amos and Yariv, Lior and Haim, Niv and Atzmon, Matan and Lipman, Yaron},
 booktitle = ICML,
 title = {Implicit Geometric Regularization for Learning Shapes},
 year = {2020}
}

@inproceedings{poco,
    title = {{POCO}: {3D} Pose and Shape Estimation using Confidence},
    author = {Dwivedi, Sai Kumar and Schmid, Cordelia and Yi, Hongwei and Black, Michael J. and Tzionas, Dimitrios},
    booktitle = I3DV,
    year = {2024}
}

@inproceedings{cseke_tripathi_2025_pico,
    title     = {{PICO}: {R}econstructing {3D} People In Contact with Objects}, 
    author    = {Cseke, Alp\'{a}r and Tripathi, Shashank and Dwivedi, Sai Kumar and Lakshmipathy, Arjun and Chatterjee, Agniv and Black, Michael J. and Tzionas, Dimitrios},
    booktitle = CVPR,
    year      = {2025},
}

@inproceedings{yi2022mover,
  author       = {Hongwei Yi and
                  Chun{-}Hao P. Huang and
                  Dimitrios Tzionas and
                  Muhammed Kocabas and
                  Mohamed Hassan and
                  Siyu Tang and
                  Justus Thies and
                  Michael J. Black},
  title        = {Human-Aware Object Placement for Visual Environment Reconstruction},
  booktitle    = CVPR,
  year         = {2022},
}

@inproceedings{zhang2020phosa,
    title = {Perceiving {3D} Human-Object Spatial Arrangements from a Single Image in the Wild},
    author = {Zhang, Jason Y. and Pepose, Sam and Joo, Hanbyul and Ramanan, Deva and Malik, Jitendra and Kanazawa, Angjoo},
    booktitle = ECCV,
    year = {2020},
}

@inproceedings{tzionas2015inhand,
    title       = {{3D} Object Reconstruction from Hand-Object Interactions},
    author      = {Tzionas, Dimitrios and Gall, Juergen},
    booktitle   = ICCV,
    year        = {2015}
}

@inproceedings{panteleris2015inhand,
	title     = {{3D} Tracking of Human Hands in Interaction with Unknown Objects},
	author    = {Panteleris, Pantelis and Kyriazis, Nikolaos and Argyros, Antonis A.},
	booktitle = BMVC,
	year      = {2015}
}

@article{savva2016pigraphs,
	title 		=	{{PiGraphs}: {L}earning interaction snapshots from observations},
	author 		=	{Savva, Manolis and Chang, Angel X and Hanrahan, Pat and Fisher, Matthew and Nie{\ss}ner, Matthias},
	journal 	=	TOG,
	volume 		=	{35},
	number 		=	{4},
	pages 		=	{139:1--139:12},
	year 		=	{2016}
}

@inproceedings{chen2019holistic++,
	title		=	{Holistic++ Scene Understanding: {S}ingle-view {3D} Holistic Scene Parsing and Human Pose Estimation with Human-Object Interaction and Physical Commonsense},
	author		=	{Chen, Yixin and Huang, Siyuan and Yuan, Tao and Zhu, Yixin and Qi, Siyuan and Zhu, Song-Chun},
	booktitle	=	ICCV,
	year			=	{2019}
}

@inproceedings{gupta2011workspace,
	title 		=	{From {3D} scene geometry to human workspace},
	author 		=	{Gupta, Abhinav and Satkin, Scott and Efros, Alexei A and Hebert, Martial},
	booktitle 	=	CVPR,
	year 		=	{2011}
}

@article{iMapper2018,
	title 		=	{{iMapper}: interaction-guided Scene Mapping from Monocular Videos},
	author 		=	{Monszpart, Aron and Guerrero, Paul and Ceylan, Duygu and Yumer, Ersin and Mitra, Niloy J},
	journal 	=	TOG,
	volume 		=	{38},
	number 		=	{4},
	pages 		=	{92:1--92:15},
	year 		=	{2019}
}

@inproceedings{kjellstrom2010trackingPeopleObjects,
  author       = {Hedvig Kjellstr{\"{o}}m and
                  Danica Kragic and
                  Michael J. Black},
  title        = {Tracking people interacting with objects},
  booktitle    = CVPR,
  year         = {2010},
}
}

\clearpage
\twocolumn[{%
\renewcommand\twocolumn[1][]{#1}%
\begin{center}
{\Large\bfseries
\textcolor{cvprblue}{RHINO}:
\textcolor{cvprblue}{R}econstructing
\textcolor{cvprblue}{H}uman
\textcolor{cvprblue}{I}nteractions with
\textcolor{cvprblue}{N}ovel
Objects \\
from Monocular Videos\par}
\vspace{0.5em}
{\Large Supplementary Material\par}
\vspace{1em}
\end{center}
}]
\renewcommand{\thesection}{S.\arabic{section}}
\renewcommand{\thefigure}{S.\arabic{figure}}
\renewcommand{\thetable}{S.\arabic{table}}
\renewcommand{\theequation}{S.\arabic{equation}}
\renewcommand{\theHsection}{supp.\arabic{section}}
\renewcommand{\theHsubsection}{supp.\arabic{section}.\arabic{subsection}}
\renewcommand{\theHsubsubsection}{supp.\arabic{section}.\arabic{subsection}.\arabic{subsubsection}}
\renewcommand{\theHfigure}{supp.\arabic{figure}}
\renewcommand{\theHtable}{supp.\arabic{table}}
\renewcommand{\theHequation}{supp.\arabic{equation}}
\setcounter{section}{0}
\setcounter{figure}{0}
\setcounter{table}{0}
\setcounter{equation}{0}

We discuss data preprocessing (\cref{supp_sec:data_preprocessing}) and implementation details (\cref{supp_sec:impl_details}), and present additional experimental results (\cref{supp_sec:exp}), including failure cases (\cref{supp_sec:failure_cases}).
Last, we discuss the possible negative societal impact (\cref{supp_sec:impact}).

\section{Data Preprocessing}
\label{supp_sec:data_preprocessing}

We adopt the data preprocessing framework of \HSR~\cite{xue2024hsr} to register human bodies and scene cameras into a common world frame. 
We extend this by adding components for: 
(i) object pose estimation, 
(ii) camera-object motion disentanglement, and 
(iii) contact estimation.
The motion disentanglement part has been detailed in the main paper \mbox{(Sec.~\textcolor{cvprblue}{3.2})}.
Here we provide more details regarding the other two parts.

\zheading{Implementation Details} %
For motion disentanglement (Sec.~\textcolor{cvprblue}{3.2}), \RANSAC automatically identifies which frames have a stationary object, without assuming that any particular frames (\eg, the first few ones) are static.
For human initialization, the weak-perspective camera used for the initial body estimation via \aios~\cite{sun2024aios} applies \emph{only} to that initialization step; we subsequently recover the body trajectory under a full perspective camera model by minimizing a 2D keypoint reprojection error.

\subsection{3D Object Pose Estimation}

\begin{figure}[t]
    \centering
    \captionsetup{type=figure}%
        \includegraphics[width=0.48 \textwidth]{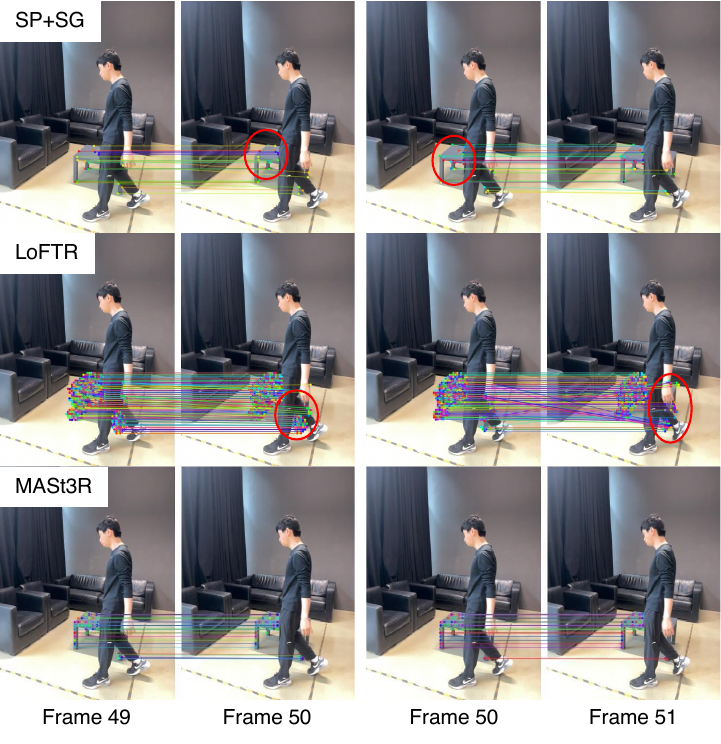}
    \vspace{-2.0 em}
    \caption{
        \textbf{Comparison of feature matches} on object pixels across consecutive frames. 
        We compare our {\tt \MASTER}-based \cite{mast3r} strategy against established baselines. 
        Red circles indicate failure modes in baselines: sparse, inconsistent keypoints ({\tt SP+SG}~\cite{detone18superpoint,sarlin20superglue}) and incorrect background matches ({\tt \LOFTR}~\cite{sun2021loftr}).
    }
    \label{supp_fig:matching}
    \vspace{-0.5 em}
\end{figure}

We detect object masks via Grounded-\SAM~\cite{ren2024grounding, ravi2025sam2} using text prompts describing the object class. 
To extract features only on object pixels, we suppress the background by replacing non-object pixels with either black or white values, dynamically chosen to maximize contrast against the object's original color. 
These masked images are then processed by \MASTER~\cite{mast3r} to extract feature matches. 
We compare our matching strategy against baselines in ~\cref{supp_fig:matching}. 
As shown in the first row, \superpoint~\cite{detone18superpoint} keypoints exhibit poor repeatability, even on identical frames.
Similarly, \loftr~\cite{sun2021loftr} yields incorrect matches outside object boundaries due to its coarse-to-fine matching strategy. 
In contrast, \MASTER yields dense and precise matches, robustifying pose estimation, and by extension shape reconstruction.

\subsection{3D Contact Estimation}
We detect 3D contact points on the human via \InteractVLM \cite{dwivedi_interactvlm_2025}, a \sota image-based model. 
However, \InteractVLM often struggles. 
First, it operates on individual frames, so there is jitter in detected 3D contact points, while neighboring frames are inconsistently labeled as ``contact" or ``non-contact" ones. 
Second, there might be false positives; the model can detect contact when there is none in the image.

We mitigate these errors via a motion-based strategy. 
Our key insight is that an object typically moves only when manipulated by a human. 
Thus, we automatically label a frame as a ``contact frame" if the temporal change in 3D object pose exceeds a threshold, else we label this as ``non-contact frame." 
However, noisy object pose estimates can cause these labels to fluctuate across frames. We tackle this by applying a temporal filter that removes short label spans by flipping them to match the surrounding label, and then extends each contact region by a margin of frames to account for the contact-but-stationary phase (\ie, when the human is in contact with the object but not yet moving it).
Then, we apply \InteractVLM only for contact frames, avoiding false positives for non-contact frames.

Finally, we remove jitter in the predicted 3D contact points by averaging per-vertex contact predictions over a local temporal window and thresholding the result.

\section{Implementation Details}
\label{supp_sec:impl_details}

\subsection{Neural Representations}

\hspace{\parindent}
\zheading{Human} 
The canonical human shape network, $f_\text{sdf}^{H}$, comprises 8 blocks, each consisting of a 256-unit fully-connected layer with weight normalization and Softplus activation. 
The human pose condition, $\boldsymbol{\theta}_b$, is a 69-dimensional vector formed by concatenating the axis-angle representations of all body joints. 
The appearance network, $f_\text{rgb}^{H}$, comprises 4 similar blocks but uses \mbox{ReLU} activation for the hidden layers and a Sigmoid for the output layer. 
To help convergence, we pretrain the shape network using a rough-shape loss using a \smplX~\cite{SMPL-X:2019} mesh in canonical pose.

\zheading{Object \& Scene} 
The shape and texture networks for the object ($f_\text{sdf}^{O}$, $f_\text{rgb}^{O}$) and the scene ($f_\text{sdf}^{S}$, $f_\text{rgb}^{S}$) share a similar architecture to the human networks. 
We employ geometric initialization following IGR~\cite{icml2020_igr}; 
the object canonical shape is initialized as a small sphere with outward-facing normals. 
Similarly, the scene geometry is initialized as a large sphere with inward-facing normals to enclose the capture volume.

\subsection{Losses}
\label{supp_sec:losses}

Following \HSR~\cite{xue2024hsr}, we train our neural representations using a combination of RGB, normal, depth, and mask losses.
However, we go beyond \HSR by modeling dynamic human-object manipulation, which causes severe occlusions, while also requiring high-fidelity hand reconstruction.
We tackle these challenges via two additional losses, $L_{\text{body}}$ and $L_\text{hand}$ described below, the effect of which is analyzed in \cref{supp_sec:exp}.

\zheading{Body Prior Loss} 
Occlusions can cause the body geometry to appear truncated. 
To resolve this, we sample points $x_\text{b}$ within the interior of the canonical \smplX~\cite{SMPL-X:2019} mesh, and penalize for predicted positive signed-distance values: 
\begin{equation}
    L_{\text{body}} = \gamma_1 \tanh \left(f_\text{sdf}^H(x_\text{b}) / \gamma_2 \right)^2 
    \quad 
    \text{for~} f_\text{sdf}^H(x_\text{b}) \ge 0
    \text{.} 
\end{equation}

\begin{figure*}[ht]
    \centering
    \captionsetup{type=figure}%
        \includegraphics[width=0.96 \textwidth]{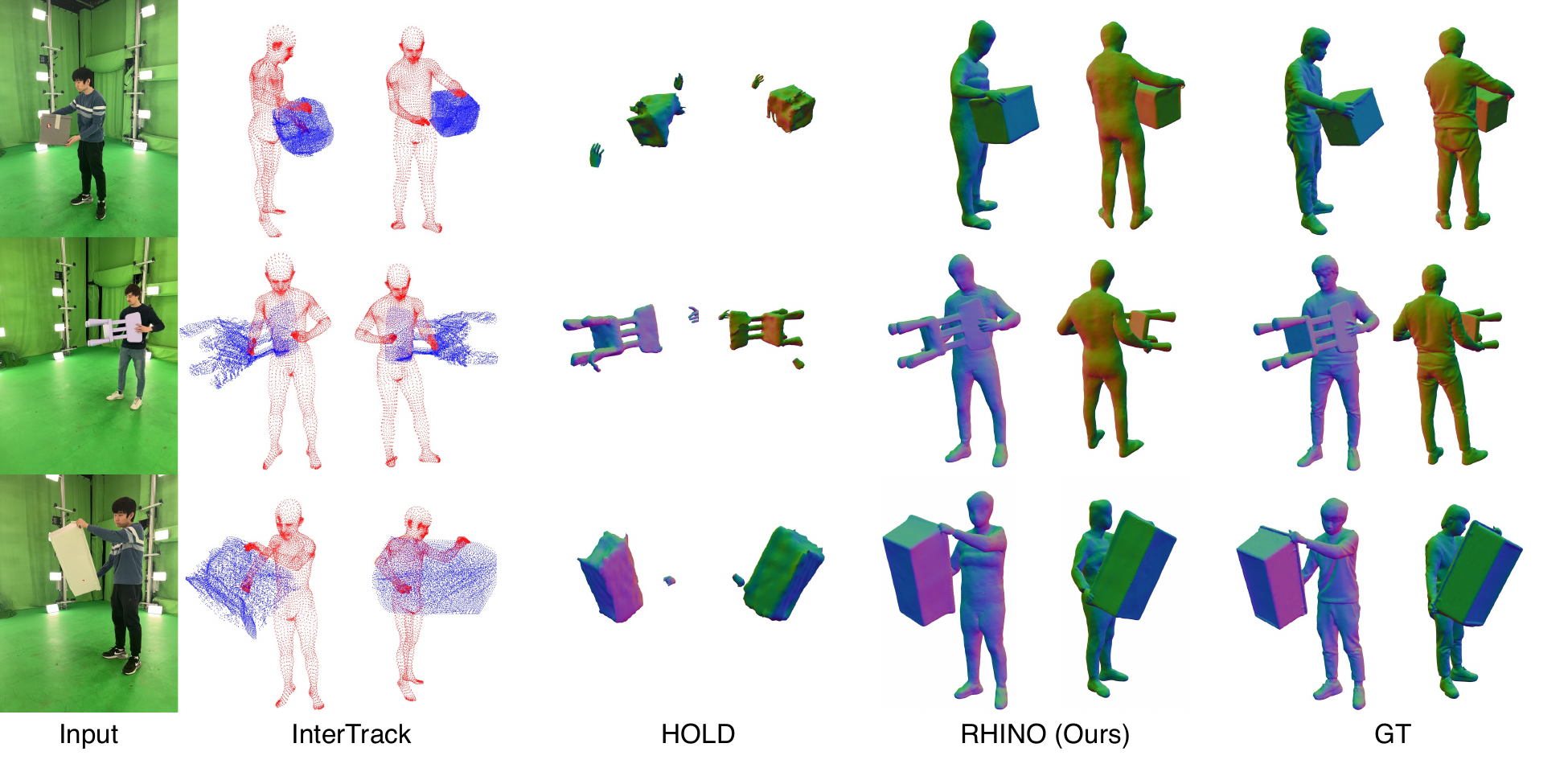}
        \vspace{-0.5em}
    \caption{
        \textbf{Evaluation on shape reconstruction (extending Fig.~\textcolor{cvprblue}{6})} using our \nameDatasetGT dataset (Sec.~\textcolor{cvprblue}{4.1}). 
        \HOLD~\cite{fan2024hold}
        struggles to reconstruct shapes under noisy object poses and fails to capture 3D hand-object interactions.
        \InterTrack~\cite{xie2025intertrack} often recovers reasonable object shape, but fails to accurately model the interaction due to errors in human and object pose or poor generalization to unseen objects. 
        Our method (\nameMethod) faithfully reconstructs detailed interactions and object shapes that align closely with the ground truth (GT).
    }
    \label{supp_fig:geo_comp}
\end{figure*}

\zheading{Hand \SDF Loss} 
The recovered hand shape can often be rough due to complex articulation and noisy pose estimation.
We resolve these by supervising the human \SDF in the hand region using the \smplX mesh as a geometric prior.
Specifically, we sample 3D points $x_\text{h}$ near the hands, compute their \SDF values $\xi(x_\text{h})$ from the \smplX mesh, and penalize the $L1$ difference of the respective prediction:
\begin{equation}
    L_\text{hand} = w(x_\text{h})| f_\text{sdf}^H(x_\text{h}) - \xi(x_\text{h})|.
\end{equation}
The weight term $w(x_\text{h})$ attenuates supervision near the wrist region, where the \smplX mesh cannot model sleeves.

\subsection{Training Details}

\hspace{\parindent}
\zheading{Alternating Optimization}
We employ a two-stage optimization strategy. 
In \mbox{Stage~I}, we optimize for all network parameters with disabled physical losses \mbox{(Eq.~\textcolor{cvprblue}{12}--\textcolor{cvprblue}{13})}. 
In \mbox{Stage~II}, we freeze the shape and appearance network parameters and optimize only for human and object poses with physical losses enabled.
We follow \mbox{Stage~I} exclusively for the first 25\% of training iterations to reconstruct initial geometry and appearance. 
Subsequently, we alternate between the two stages every 10 epochs (6 epochs for \mbox{Stage~I} and 4 epochs for \mbox{Stage~II}). 
The model is trained for 100k steps, taking $\sim$18 hours on an \mbox{NVIDIA} RTX \mbox{4090} GPU.

\zheading{Sampling Strategy}
We use a weighted pixel sampling strategy, allocating 50\% of samples to the human body, 30\% to the object, 10\% to the hand, and 10\% to the scene. 
To help reconstruct the initial object shape, for the first 10 epochs we increase the object sampling rate to 70\%, while reducing body samples to 20\% and hand samples to 0\%.

\subsection{Evaluation Details}

Different methods produce output in different formats (\eg, meshes or point clouds, presence/absence of human or object) and operate within different coordinate frames. 
Thus, we tailor our evaluation protocols for fair comparisons.

\zheading{Coordinate Frames}
\HOLD~\cite{fan2024hold} and \InterTrack~\cite{xie2025intertrack} produce output in the camera frame, so post hoc we need to align it into the world frame. 
Instead, \HSR~\cite{xue2024hsr} and \nameMethod operate in the world frame, but still require rigid alignment to account for differences in camera trajectory.

\zheading{Aligning: Estimated Shapes to GT Shapes}
To compute evaluation metrics, we first need to align the estimated shapes to the GT ones, as follows below for each method:
\begin{itemize}
\item
\HSR estimates clothed human meshes, and does not estimate moving objects. 
Therefore, we perform frame-wise alignment using the ground-truth (GT) human mesh.
\item
\HOLD and its variant \HOLDplusRHINO (that uses \nameMethod's object pose estimates), estimate hands and a manipulated object, focusing on object reconstruction quality. 
Thus, we align the predicted object mesh with the GT one.
\item
\InterTrack estimates unclothed human bodies as point clouds (\SMPL vertices) rather than clothed meshes as in the GT, so direct surface alignment is infeasible. 
Instead, we perform Procrustes alignment on 3D joints derived from the \InterTrack predictions and GT \smplX fits. 
We perform only human-based alignment, as \InterTrack's object poses are often noisy, with wrong scale or orientation. 
\item
\nameMethod estimates both the human and object, so we perform alignment based on the human, the object, or both. 
\end{itemize}

\zheading{Results}
In Tab.~\textcolor{cvprblue}{1} of the main manuscript, we report metrics after aligning the estimated shapes of each method to GT ones appropriately, as described above.

The \nameDatasetGT dataset contains 7 sequences: \texttt{S1\_Take3}, \texttt{S1\_Take7}, \texttt{S1\_Take8}, \texttt{S2\_Take5}, \texttt{S2\_Take9}, \texttt{S3\_Take6}, and \texttt{S4\_Take2}.
In Tab.~\textcolor{cvprblue}{3}, the ``\mbox{{\tt SP+SG}}'' baseline yields plausible object shapes on only 5 of these, while ``{\tt \loftr}'' succeeds on 6.
For fair comparison, we report aggregated metrics on the 5 sequences where all methods succeed: \texttt{S1\_Take3}, \texttt{S1\_Take7}, \texttt{S2\_Take5}, \texttt{S2\_Take9}, and \texttt{S3\_Take6} (excluding \texttt{S1\_Take8} and \texttt{S4\_Take2}).

\section{Additional Experiment Results}
\label{supp_sec:exp}

\subsection{Shape Reconstruction}

We provide more qualitative comparisons on shape reconstruction in ~\cref{supp_fig:geo_comp}. 
The results reinforce our observations regarding the limitations of baseline methods for challenging object shape and noisy pose initialization.
\HOLD fails to generate coherent object meshes when the initial object pose is noisy (rows~1,~3 of \cref{supp_fig:geo_comp}), while the recovered hands lie far away from the ``grasped'' object. 
\InterTrack struggles with generalization, particularly with the white box (row~3). 
Furthermore, \InterTrack misaligns the object relative to the human, failing to preserve the correct spatial configuration required for a valid grasp (rows~1,~2).
In contrast, our method (\nameMethod) faithfully reconstructs detailed object shapes and maintains better hand-object contact.

\subsection{Novel--View Synthesis}

\begin{figure}
    \centering
        \vspace{-0.5 em}
    \captionsetup{type=figure}%
        \includegraphics[width=0.47 \textwidth]{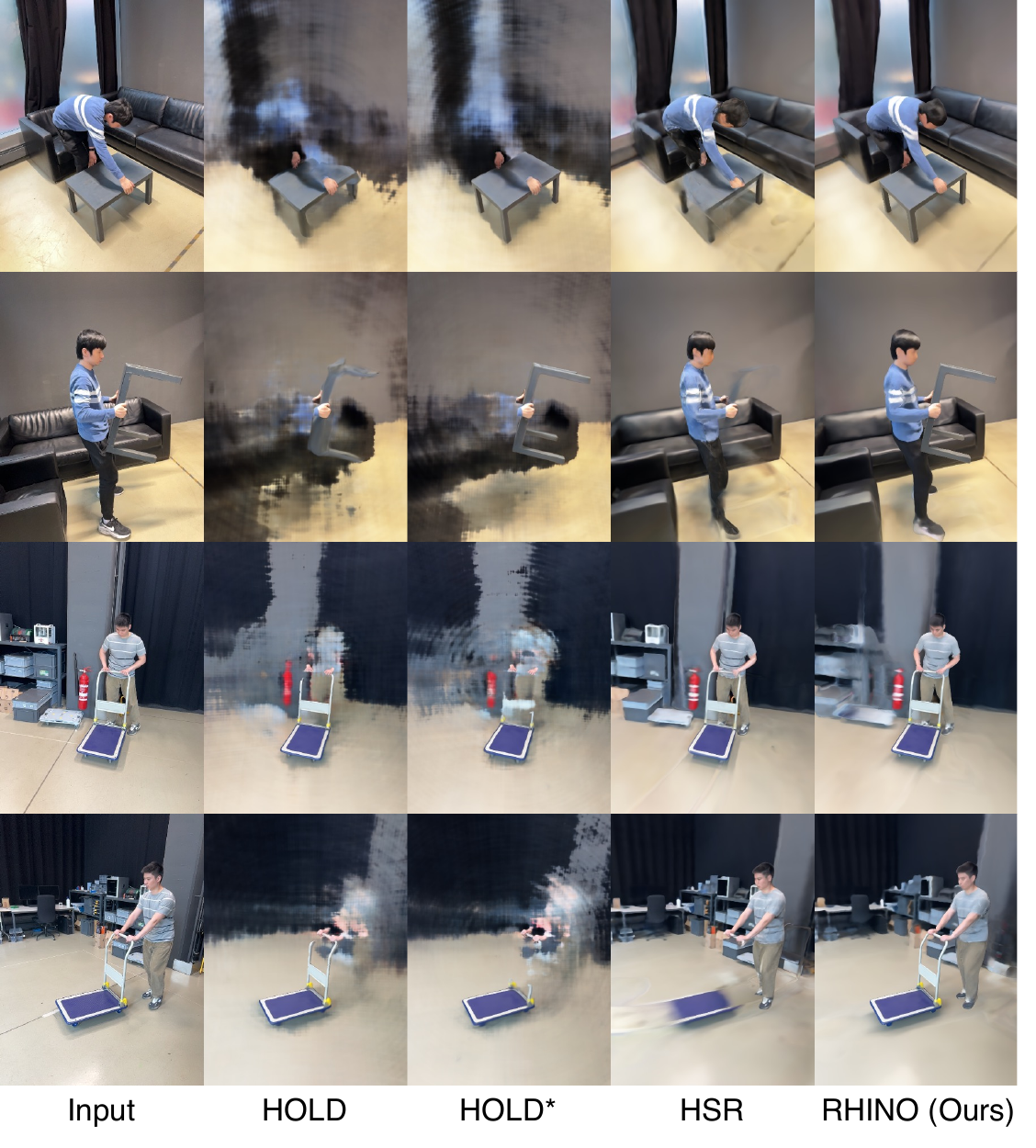}
        \vspace{-1.0 em}
    \caption{
        \textbf{Evaluation on novel view synthesis} on \nameDatasetITW.
        \HOLD~\cite{fan2024hold} fails to reconstruct the background scene due to operating in the camera frame.
        \HSR~\cite{xue2024hsr} cannot capture dynamic manipulated objects as it assumes a static scene. 
        Our \nameMethod framework decouples the dynamic foreground from the background, and yields high-fidelity renderings 
        that closely resemble the input. 
    }
    \label{supp_fig:nvs}
\end{figure}

We evaluate qualitatively on novel view synthesis using the \nameDatasetITW dataset. 
The results in \cref{supp_fig:nvs} show that baselines have distinct failure modes, which \nameMethod successfully overcomes. 
\HOLD~\cite{fan2024hold} struggles with background reconstruction (see column 2). 
This is because \HOLD operates in camera frame, so it lacks a global understanding of scene geometry. 
Thus, as the camera viewpoint changes, the background appears fragmented and noisy, failing to maintain temporal or spatial consistency. 
\HSR~\cite{xue2024hsr} is capable of modeling the static background, but struggles significantly with the dynamic foreground elements; 
the rendered images suffer from ``ghost" artifacts around the object (col.~3). 
Note in rows~1 and 3 that, although the object appears well-reconstructed, it is baked into the static background scene rather than being correctly modeled as a dynamic entity. 
In contrast, our \nameMethod framework yields high-fidelity renderings that differ negligibly from the input observations. 
By effectively decoupling the dynamic foreground from the static background, it preserves the structural integrity of the object even during fast motion or complex interactions.

\begin{figure}
    \centering
        \vspace{-0.5 em}
    \captionsetup{type=figure}%
        \includegraphics[width=0.49 \textwidth]{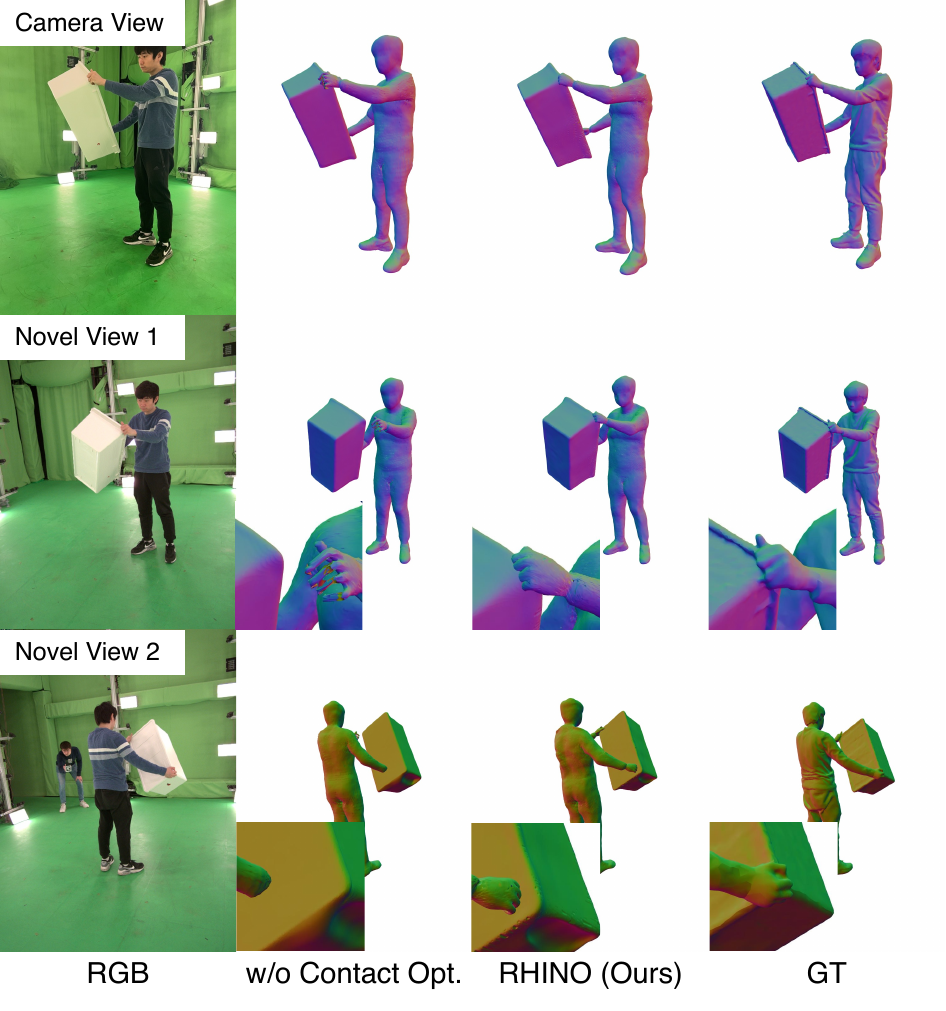}
        \vspace{-2.40 em}
    \caption{
        \textbf{Ablation on the effect of contact} on \nameDatasetGT. 
        Without contact optimization (``{\tt w/o~Contact Opt.}''), the reconstructed 
        hands penetrate the object, or hover and fail to contact it (see \mbox{zoomed-in} insets). 
        In contrast, our full method (``{\tt \nameMethod~(Ours)}'') improves physical plausibility, recovering grasps that better align with the ground truth ({\tt ``GT''}). 
    }
    \label{supp_fig:contact}
\end{figure}

\subsection{Ablation on Contact}

We evaluate our contact-based pose refinement in \cref{supp_fig:contact}. 
We compare our full framework (``{\tt \nameMethod (Ours)}'') against an ablated version where the physical contact losses (Eq.~\textcolor{cvprblue}{12}--\textcolor{cvprblue}{13}) are removed (``{\tt w/o~Contact Opt}'').

Relying solely on visual cues often results in depth ambiguities and physical implausibility, particularly under heavy occlusion. 
While the ablated model (``{\tt w/o~Contact Opt}'') may achieve good alignment in camera view, novel-view renderings have strong artifacts. 
Specifically, as shown in the \mbox{zoomed-in} insets of \cref{supp_fig:contact}, the hands often penetrate the object volume or hover above the surface. 
In contrast, our full model (``{\tt \nameMethod (Ours)}'') successfully resolves these ambiguities, producing tight, physically plausible grasps that closely match the ground truth (``{\tt GT}").

\subsection{Failure Modes}
\label{supp_sec:failure_cases}

We identify the following typical failure modes of \nameMethod:

\zheading{Insufficient Object Coverage}
When the object is observed from a limited range of viewpoints (\eg, always from the front), the \SFM-based pose initialization may lead to incomplete or inaccurate object shape on unseen sides.

\zheading{Rapid Motion}
Rapid object or hand motion can cause severe motion blur and large inter-frame displacements, making feature matching unreliable and causing the motion decomposition (Eq.~4) to yield noisy object trajectories.

\zheading{Extreme Occlusion}
When the object is almost entirely occluded by the body for many consecutive frames, the object's appearance and pose are under-constrained. The neural SDF may hallucinate geometry in unobserved regions.

\zheading{Non-Rigid Objects}
We assume rigid objects. For deformable objects (\eg, cloth, soft toys), the rigid-body motion model is insufficient, causing reconstruction artifacts.

These failure modes inform future work, toward handling articulated or deformable objects, leveraging temporal priors for rapid motions, and generative shape priors for occlusions.

\section{Societal Impact}
\label{supp_sec:impact}

\nameMethod converts humans and objects into digital forms from a single \RGB video, unlocking vast possibilities for augmented and virtual reality, assistive robotics, and learning from internet-scale videos. 
Our framework generates \HOIs that can be animated to previously unseen poses. 

\zheading{Risks \& Mitigation}
The above capability carries an inherent risk of misuse, most notably for creating deep-fakes. 
It is crucial to address such concerns before this technology is integrated into products. 
We are committed to fostering applications that provide a clear benefit to society. 
While we cannot eliminate the possibility of malicious use, we advocate for a policy of maximum transparency. 
To this end, openly discussing our methods and sharing both code and data is not only an ethical imperative but also a practical way to enable the development of countermeasures, thus mitigating the dangers of harmful applications.

\end{document}